\documentclass[preprint,12pt,3p]{elsarticle}

\usepackage{amssymb}
\usepackage{amsmath}
\usepackage{booktabs}
\usepackage{makecell}
\usepackage{url}
\usepackage[authoryear]{natbib}
\usepackage{hyperref}
\usepackage{subfig}

\hypersetup{
    colorlinks=true,
    linkcolor=blue,
    filecolor=blue,      
    urlcolor=blue,
    citecolor=blue,
}

\journal{Elsevier}

\begin{document}

\begin{frontmatter}

\title{On Forecast Stability}

\author[1]{Rakshitha Godahewa}
\author[2,3,1]{Christoph Bergmeir \corref{cor1}}
\author[4]{Zeynep Erkin Baz}
\author[4]{Chengjun Zhu}
\author[1]{Zhangdi Song} 
\author[2,3]{Salvador Garc\'ia}
\author[4]{Dario Benavides}

\address[1]{Department of Data Science and Artificial Intelligence\\
    Monash University, Australia.}
\address[2]{Department of Computer Science and Artificial Intelligence\\
	University of Granada, Spain.}
\address[3]{Data Science and Computational Intelligence Andalusian Institute, DaSCI, Andalucía, Spain.}
\address[4]{Meta Inc., California, USA}

\cortext[cor1]{Corresponding Author: bergmeir@ugr.es}

\begin{abstract}
Forecasts are typically produced in a business context on a regular basis to make downstream decisions. Here, forecasts should not only be as accurate as possible, but also should not change arbitrarily, and be stable in some sense. In this paper, we explore two types of forecast stability that we call vertical stability (for forecasts from different origins for the same target) and horizontal stability (for forecasts from the same origin for different targets). Existing works in the literature are only applicable to certain base models and can only stabilise forecasts vertically. We propose a simple linear-interpolation-based approach to stabilise the forecasts provided by any base model, both vertically and horizontally. Our method makes the trade-off between stability and accuracy explicit, producing forecasts at any point in the spectrum of this trade-off. Using N-BEATS, Pooled Regression, LightGBM, ETS, and ARIMA as the base models, in our evaluation across different error and stability measures on four publicly available datasets, on some datasets, the proposed framework achieves forecasts that are both more accurate and stable than the base forecasts. On the others, we achieve forecasts that are slightly less accurate but much more stable.
\end{abstract}

\end{frontmatter}

\section{Introduction}
\label{sec:intro}

In many business applications, forecasts are produced on a regular rolling basis, and for a window of target future dates. If the forecasts are volatile that can have negative consequences for subsequent decision-making steps. Thus, the stability of forecasts of some sort is often a desirable property. Stability can be understood in different ways. For example, it can mean that forecasts performed on different (origin) days for the same target day (different forecast \emph{revisions}) should not differ too much, or it can mean that forecasts within an output window should be stable/smooth in some way. It can also simply mean an ensembling/forecast combination step to ``stabilise'' the output of low-bias high-variance forecasting methods, such as neural networks. Finally, it can indicate any combination of these factors. 

There is often a perceived trade-off between the stability and accuracy of the forecasts. For example, in the case of forecasts from different origins for the same target, as new information becomes available, it is assumed that the updated forecasts will be more accurate by incorporating this new data. However, \cite{pritularga2024forecast} argue that accuracy and stability are only weakly correlated. They use the term ``congruence'' as they argue that stability is a reserved term in statistics, and define forecast congruence as the variance over forecasts from different origins for the same target. Their theoretical framework shows that accuracy and congruence are based on the same raw errors, differing only in summarisation. 
Thus, both can be improved simultaneously up to a point, after which improving one may worsen the other. 
Their experiments in a supply chain application demonstrate that achieving acceptable accuracy with good congruence is often more related to good downstream decision-making outcomes than achieving the best accuracy alone. They identify methods using shrinkage and temporal hierarchies as inherently producing more congruent forecasts.
In earlier work, \citet{VANBELLE20231333} follow more explicitly the approach of building a model that is inherently able to produce forecasts that are stable and have competitive accuracy. They propose modifying the loss function of the original N-BEATS framework \citep{oreshkin_2019_nbeats} to optimise both forecast accuracy and stability.
Using a dedicated model has advantages, for example that the model is also optimised for stability, not only accuracy, but also a couple of drawbacks that may arise in certain real-world situations.
Stability is built into the models at training time, and stabilisation is performed not against the actual prior forecasts that are available at inference time, but against implicitly built ``prior'' forecasts during training, with the goal to achieve a model that on average obtains stable forecasts over many origins. Thus, there is little that can be changed at inference time. In practice, we may have forecast-value-added pipelines where forecasts are adjusted judgementally, and stability in practice will be the volatility of the final forecast after the adjustments. Also, the forecasting methodology could change between different forecasts, or bespoke models that do not consider stability may be deployed in production and changing them or their loss function may be difficult.

Therefore, in this paper, we adopt a different approach. Instead of proposing a dedicated forecasting method that can produce forecasts that inherently trade off accuracy and stability in some optimal way, we take a step back and propose a simple and generic yet powerful post-processing approach using linear interpolation. Although it may not yield an optimal trade-off, as there is no explicit optimisation of stability, it has certain advantages. It is model-agnostic and can work with any base model, particularly in data-scarce situations, such as with local models or judgemental forecasts. Thus, it can be easily integrated into existing pipelines and workflows. Furthermore, it allows us to flexibly manage the trade-off between accuracy and stability, without re-fitting the base models.
As the approach is straightforward, it has been implicitly used in some forms in the literature. \citet{VANBELLE20231333} use a comparison method that employs the N-BEATS model, which they call N-BEATS origin ensemble. For the final forecast of a particular target time point, this method produces a simple average of the forecasts made at prior origins and the current origin corresponding to that target time point. 
The winning approach of the M5 forecasting competition \citep{IN20221386} also implicitly considers the simple average of the forecasts obtained using different forecast origins as the final forecasts. 
Thus, the forecasts are stabilised and as the method won the competition, the forecasts are highly accurate as well. However, those authors did not consider their work from a stability aspect, and did not explore the stability of the forecasts provided by the method.

Beyond the methodological contribution, we also categorise different types of forecast stability, drawing from the supply chain planning literature but focusing on forecasting rather than the subsequent planning stage. Our experiments show that our approach often produces forecasts that are both more accurate and stable than the base models. In other cases, it yields significantly more stable forecasts, with only modest losses in accuracy. All implementations of this study are publicly available at: \url{https://github.com/rakshitha123/StableForecasting}. 

In summary, our study is the first to present a systematic analysis showing the use of linear interpolation to stabilise forecasts using state-of-the-art machine learning tools and datasets. We apply linear interpolation to the forecasts made at different forecast origins, separately, considering both simple and weighted averaging. It has the advantage that it can be used in a model-agnostic manner, even for situations where different methodologies are used for different forecast iterations (i.e., different origins), it is fast to compute, and it allows in a very straightforward way to control the trade-off between accuracy and stability, without the need to recompute the base forecasts.

The remainder of this paper is organised as follows. Section \ref{sec:related_work} discusses further related work.
Section \ref{sec:methodology} introduces our stability categorisation and our proposed linear interpolation approach in detail. Section \ref{sec:experiments} discusses the experimental framework, including datasets, error measures, base models, and benchmarks. Section \ref{sec:eval_accuracy} presents the analysis of the results. Section \ref{sec:conclusion} concludes the paper and discusses possible future research directions.

\section{Related Work}
\label{sec:related_work}

Apart from the work already discussed in Section~\ref{sec:intro}, there is limited literature that explicitly considers the stability of forecasts. \cite{chae2009developing} proposes a set of KPIs, including `forecast volatility,' which measures changes in forecasts across revisions. 
\cite{schuster_2017_bridging} consider a planning exercise in the semi-conductor industry with demand forecasts for different horizons and granularities. They use the terms inter-plan stability for stability across forecasting revisions, and intra-plan stability for forecasts from the same origin within an output window.

Much of the existing literature considers stability in a supply chain context, where forecast revisions can lead to costly changes in supply chain plans. While these works are related to our work, the difference is that the works in this research field usually do not consider the forecasts explicitly and how they could be made more stable, but rather consider the consequences of given forecasts on the stability of a subsequent planning phase. 
In this context, changing forecast revisions lead to revisions of supply chain plans and can incur significant costs for businesses \citep{li_2017_revisiting,tunc_2013_simple,chae2009developing}. 
Making the forecast stable (without loosing too much accuracy), in the sense of producing a smooth forecast over an output window, can also be important in this application area to reduce the bullwhip effect \citep{lee_1997_bullwhip,wang2016bullwhip}. This effect refers to the phenomenon of demand variation amplification in a supply chain consisting of a large number of parties, including manufacturers, wholesalers, suppliers and customers. Unstable forecasts corresponding to one party in a supply chain may lead to higher fluctuations in demand information for other parties, incurring significant costs. 
Many studies have addressed the instabilities in material requirement plans (MRP), mostly caused by unstable demand forecasts. We discuss some examples of works in this area in the following. In this context, instability is often called ``nervousness'', a review can be found in \cite{sahin2013rolling}.  \cite{herrera2009simulation} present a method for making a plan significantly more stable at only a slightly higher cost. Those authors refer to changes between forecast revisions as `nervousness', and to changes within the output window of the same forecast as `instability.'
In a recent paper, \cite{saez2023reducing} develop an algorithm with distributed agents to reduce planning nervousness. They call different forecast revisions `cycles' and output windows `periods'. Those authors then examine nervousness both across periods and cycles.
Considering forecasting more explicitly, \cite{dejonckheere2002transfer} analyse the exponential smoothing model and how its smoothing parameters can be selected to make subsequent plans less susceptible to instabilities in the form of the bullwhip effect. \cite{li2014avoiding} perform a similar exercise, incorporating a damped trend into their model.

Ensembling is another approach to mitigating forecast instability. In the forecasting space, ensembling is also known as forecast combination \citep{granger_1969_combination,ref_58}. Ensembled models aggregate the predictions provided by multiple models to obtain the final predictions. Ensembling can reduce model variance and bias \citep{schapire_1999_adaboost,breiman_2001_random} and thus, provide stable forecasts compared to individual forecasting models \citep{kolassa_2011_combining, Yuan_2005_combining}. Forecast combinations are also used to stabilise forecasts in the macroeconomic forecasting domain \citep{altavilla_2007_information}.

\section{Methodology}
\label{sec:methodology}

In this section, we first explain the proposed stability categorisation, and then introduce the proposed model-agnostic approach that can be used to stabilise forecasts.

\subsection{Categorisation of Types of Forecast Stability}
\label{sec: stability_types}

Completing categorisations in the literature from a supply chain planning context \cite[see, for example,][]{schuster_2017_bridging}, we identify four different types of stability. These depend on the same or different forecast origins and targets of the methods, as shown in Figure \ref{fig:stable_intro}. In particular, we may want to achieve stability between the forecasts that have been produced:

\begin{figure}
\centering
  \includegraphics[width=0.9\textwidth]{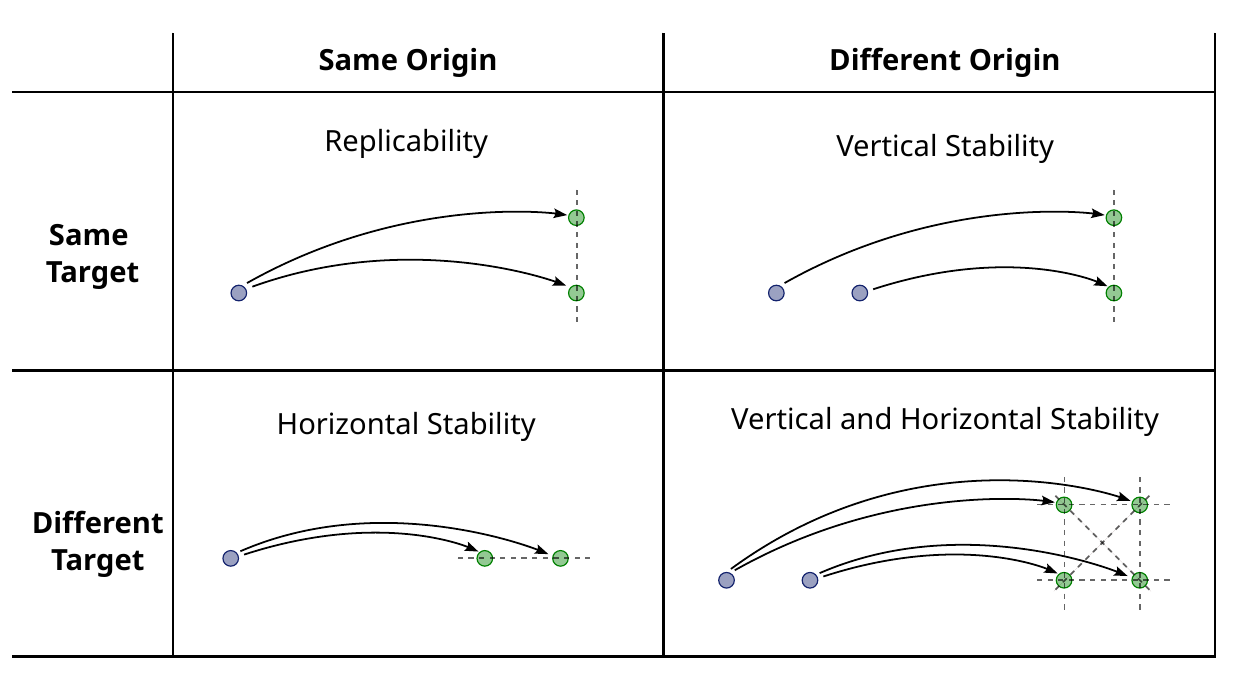}
  \caption{Forecasts can be produced from the same or different origins, and for the same or different target. We call the four different possible combinations accordingly: replicability, vertical stability, horizontal stability, vertical and horizontal stability. The blue and green dots respectively represent the origin that the forecasts are made and the target.}
  \label{fig:stable_intro}
\end{figure}

\begin{enumerate}
\item from the same origin and for the same target. This is effectively ensembling or a forecast combination approach and can also be called \emph{replicability}.
\item from different origins for the same target. 
We call this \emph{vertical stability} in the following, as the forecasting target is at the same time stamp, and therewith the desired stability is vertical on the time axis.
This type of (in)stability has been called in the literature rolling origin stability, forecast congruence, or, in a supply chain planning context, inter-plan stability, nervousness, nervousness across cycles, or forecast volatility.

\item from the same origin for different targets. 
We call this \emph{horizontal stability}, as the desired stability is horizontal with respect to the time axis of the forecast targets. In a supply chain planning context, this type of (in)stability has been called intra-plan stability, instability, or nervousness across periods.
\item from different origins for different targets. This form of stability is a mix of both horizontal and vertical stability.
Here, the desired stability could be either sequentially first vertically then horizontally or vice versa, or we could aim to achieve stability directly between forecasts that have both different origins and targets.

\end{enumerate}

Ensembling has been extensively covered elsewhere and it is not the main focus of our paper. Thus, we do not discuss it further in the paper and focus in the remainder of the paper on vertical and horizontal stability.

\begin{figure}
\centering
  \includegraphics[scale=0.7]{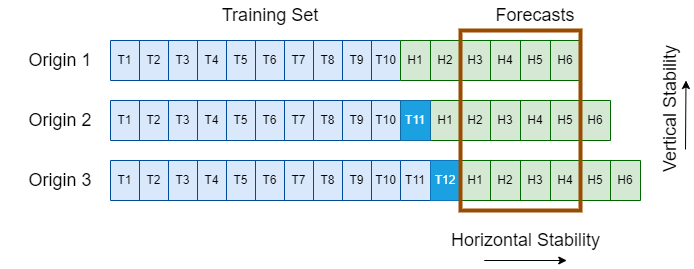}
  \caption{Visualisation of the concepts of vertical and horizontal stability types.}
  \label{fig:vertical_concept}
\end{figure}

\paragraph{Vertical Stability} In real-world applications, forecasts are often obtained in a rolling origin fashion. Thus, forecasts corresponding to a particular target time point are obtained multiple times considering different forecast origins as shown in 
Figure~\ref{fig:vertical_concept}. Let us assume a time series with 10 data points, T1-T10. A forecasting model is trained with the training data and 6-step ahead forecasts are obtained, H1-H6, with T10 as the origin. The number of forecasts made at a given origin, i.e., the forecasting horizon or forecasting window, is 6 in this example.

In real-world applications, new data points become available over time. Thus, when the next data point, T11, becomes available, it is added to the training set. Again, 6-step ahead forecasts are made considering T11 as the new, second, origin. When T12 becomes available, it is also added to the training set, and considering T12 as the third origin, 6-step ahead forecasts are again made. We can see from the figure that the forecast output windows overlap, so that forecasts made at different origins are corresponding with the same target time point. For example, consider the forecasts provided by adjacent forecast origins. Here, the forecasts H2-H6 at origin 1 and H1-H5 at origin 2, and H2-H6 of origin 2 and H1-H5 of origin 3 are corresponding with the same time periods. Non-adjacent forecast origins also provide forecasts corresponding with the same time period, e.g., forecasts H3-H6 at origin 1, and H1-H4 at origin 3 (brown box). Thus, potentially many more than two forecasts will be made for the same target value, depending on the step size in which the origin is moved and the size of the forecasting window.

To achieve vertical stability, our goal is to ensure that forecasts produced at multiple origins for the same time periods are close. 
Otherwise, decisions based on earlier forecasts (e.g., the number of products to order for the next 6 weeks) may require significant changes when new forecasts are provided later. This can lead to substantial order-adjustment costs.
Thus, for proper decision-making, forecasts corresponding to the same time periods made at different origins should be as vertically stable as possible without sacrificing (too much) accuracy. 

Only the last forecasts are produced for the time points at which no previous forecasts already exist. As such, these new forecasts can be produced purely to maximise accuracy. All other forecasts are merely updates of existing forecasts. We assume that previous forecasts have already been communicated to stakeholders. As such, they cannot be changed anymore, and the new forecasts need to be ``anchored'' at the old forecasts to achieve vertical stability. 
Assuming that newer forecasts are more accurate because they include new data and predict with a shorter horizon, we observe a potential trade-off between accuracy and stability here.

\paragraph{Horizontal Stability}

Horizontal stability is for forecasts produced from the same origin. As shown in Figure~\ref{fig:vertical_concept}, to achieve horizontal stability, the forecasts H1-H6 that are made at the same origin are required to be close to each other. Thus, the adjacent forecasts: H1 and H2, H2 and H3, H3 and H4, H4 and H5, and H5 and H6, need to be close. This eventually tries to make the forecasts smooth throughout the forecast horizon.

Horizontally stable forecasts are useful in some real-world applications to counter the so-called bullwhip effect in supply chains. For example, consider a supply chain containing a manufacturer, a supplier, and customers. When product sales at the customer's end are predicted to increase, more products must be ordered from the supplier. In general, the supplier orders products from the manufacturer with a buffer, and thus orders more products from the manufacturer to fulfill customer demand. The manufacturer also produces goods with a buffer and thus, it will produce even more goods. Thus, small fluctuations at the customer end may result in larger fluctuations at the manufacturer end.  Horizontally stable forecasts reduce the fluctuations in the forecasts from the consumer end and thus reduce the bullwhip effect.  

Note that horizontal stability is often undesirable in the strict sense. For example, when there are known future promotions in retail, we expect the future demand (and its accurate forecasts) to change sharply and not be smooth. In addition, other highly predictable patterns, namely trends and seasonality, can be present. Thus, in general, we want to develop horizontally stable forecasts that are not necessarily stabilised towards a flat line, but towards a certain seasonality and/or trend, so that they show smooth seasonalities and trends, or we could introduce smoothing weights that change based on the amount of smoothing desired. 

\paragraph{Joint Horizontal and Vertical Stability}

As horizontal and vertical stability serve different purposes, there can be situations in which both are desirable. Again, forecasts made at earlier origins would already have been communicated to stakeholders and are fixed, so that we can only stabilise the forecasts from later origins towards forecasts from earlier origins. We can do this either in a sequential manner, first vertically stabilising then horizontally stabilising or vice versa, or we could directly stabilise forecasts from different origins and horizons. For example, in Figure~\ref{fig:vertical_concept}, the forecasts H2-H5 from origin 2 could all be stabilised using the forecasts H3-H6 from origin 1, to make them both vertically and horizontally more aligned.

\subsection{Proposed Framework}
\label{sec: framework}

We propose a framework based on linear interpolation to stabilise forecasts. The methods of applying linear interpolation to obtain vertical and horizontal forecast stability are explained in the following.

\subsubsection{Linear Interpolation for Vertical Stability}
\label{sec: vert_int}

As shown in Figure \ref{fig:vertical_concept}, consider an example situation where the forecasts are required to be vertically stabilised across 3 consecutive origins where 6-step ahead forecasts are made at each origin. Thus, in this case, the forecast horizon (h) is 6. The forecasts, H2-H6 at a given origin and the forecasts, H1-H5 at the next origin correspond to the same time period. Thus, these forecasts should be close to each other in order to achieve vertical stability.

Note that the forecasts made at the first origin of each series cannot be stabilised as they are the first forecasts obtained from the series, and forecasts for later origins have not yet been generated. This is, to some degree, a consequence of our method being a post-processing method in which no implicit prior or posterior forecasts are generated within the method for the mere sake of stabilising other forecasts. The last forecast made at each origin also cannot be stabilised as these are the first forecasts corresponding to a particular time point. Thus, for each series, the forecasts are required to be stabilised from the second origin onwards, except for the last forecasts. 

The forecasts made at a given origin can be stabilised by linearly combining them with the forecasts made at the previous origin in a pairwise manner, or by linearly combining them with the forecasts made at all prior origins together. 

Thus, stable forecasts at the second origin are obtained as a linear combination of the original forecasts made at the second origin and the corresponding forecasts made at the first origin as shown in Equation \ref{eqn:origin2}. Here, $SF_{O_{2}H_{j}}$ is the stable forecast of the $j^{th}$ horizon at origin 2, $F_{O_{2}H_{j}}$ is the original forecast of the $j^{th}$ horizon made at origin 2, $F_{O_{1}H_{j+1}}$ is the corresponding original forecast made at origin 1, and $w\_s$ is the weight of the corresponding forecast made at origin 1, where $1 \leq j < h$ and $0 \leq w\_s \leq 1$. 

\begin{equation}
\label{eqn:origin2}
    SF_{O_{2}H_{j}} = (w\_s) F_{O_{1}H_{j+1}} + (1-w\_s) F_{O_{2}H_{j}}
\end{equation}

From the third origin onwards, linear interpolation can be performed in two ways to stabilise the forecasts. We refer to these two methods as \textit{partial interpolation} and \textit{full interpolation}. Equations \ref{eqn:partial} and \ref{eqn:full} show the formulas of partial and full interpolation, respectively, where $SF_{O_{i}H_{j}}$ is the stable forecast of the $j^{th}$ horizon at $i^{th}$ origin, $F_{O_{i}H_{j}}$ is the original forecast of the $j^{th}$ horizon made at $i^{th}$ origin and $w\_s$ is the weight of the corresponding forecast made at $(i-1)^{th}$ origin, where $i > 2$ and $1 \leq j < h$ and $0 \leq w\_s \leq 1$. 

\begin{equation}
\label{eqn:partial}
    SF_{O_{i}H_{j}} = (w\_s) F_{O_{i-1}H_{j+1}} + (1-w\_s) F_{O_{i}H_{j}}
\end{equation}

\begin{equation}
\label{eqn:full}
    SF_{O_{i}H_{j}} = (w\_s) SF_{O_{i-1}H_{j+1}} + (1-w\_s) F_{O_{i}H_{j}}
\end{equation}

As shown in Equations \ref{eqn:partial} and \ref{eqn:full}, the difference between partial and full interpolation methods is that to obtain stable forecasts at the current origin, partial interpolation considers the corresponding original forecasts at the previous origin whereas full interpolation considers the corresponding interpolated stable forecasts at the previous origin. Thus, partial interpolation considers the forecasts made at adjacent forecast origins and combines them in a pairwise manner. Although full interpolation combines the forecasts corresponding to adjacent forecast origins, it also considers the forecasts made at all prior origins, as the prior interpolated stable forecasts are used during the interpolation. In this sense, full interpolation combines the forecasts in a chained manner, where higher weights are given to the forecasts made at closer origins. In general, full interpolation is closer to a practical use case as the forecasts at a given origin should be stable with respect to the forecasts corresponding to the prior origin, not the original forecasts, but the stable forecasts that are already communicated to stakeholders.

\subsubsection{Linear Interpolation for Horizontal Stability}
\label{sec:hor_int}

The concept of horizontal stability considers adjacent forecasts from a given origin. For the example shown in Figure \ref{fig:vertical_concept}, it requires the adjacent forecasts such as H1 and H2, H2 and H3, H3 and H4, H4 and H5, and H5 and H6 of any origin to be closer. 

We note that for horizontal stability, it will depend on the application case in which way exactly stability is required. If the series have trends and seasonality, we may want to make them stable towards these patterns, so that the forecasts do not deviate too much from an average seasonal pattern, for example. Similar considerations can be made for external factors such as promotions. 
Another important question is whether these effects will be taken into account during the forecasting phase or later during the stabilisation phase. For the former case, we require a forecasting method that produces components, as, e.g., exponential smoothing does it. Then, stabilisation can focus on the forecasts of the remainder series. In this case, stabilisation is no longer model-agnostic, as it relies on the forecasting model to produce components. The other option is to keep the method model agnostic, in a way that during post-processing we decide on the components and then produce dedicated forecasts of these components only for the purpose of stabilisation.
This second approach makes the stabilisation method more complex and application-dependent, as a good choice of components may depend on which seasonalities and trends are present in the series. 
Thus, for simplicity and generality of the approach, we focus on the first case, where stabilisation is performed towards a flat line, either because the series have no seasonal and trend patterns or because we stabilise only the remainder component of a forecasting method that produces forecasts of components.

In an iterative forecasting scheme to generate multi-step ahead forecasts, it is not possible to horizontally stabilise the forecasts corresponding to the first horizon, H1. In a direct scheme, while it is possible to stabilise also the H1 forecast with forecasts for larger horizons, it is often a valid assumption that the forecast for H1 is the most accurate one as it has the smallest horizon. Thus, in our work, for each series and origin we stabilise the forecasts from H2 onwards, where the forecast for H1 serves as an initial anchor.

In the most generic case, when stabilising towards a flat line, the stable forecasts corresponding to H2 are obtained as a linear combination of the original H2 and H1 forecasts, as shown in Equation \ref{eqn:h2}. Here, $SF_{H_{2}}$ is the stable H2 forecast, $F_{H_{2}}$ is the original H2 forecast, $F_{H_{1}}$ is the original H1 forecast, and $w\_s$ is the weight given for the H1 forecast, where $0 \leq w\_s \leq 1$.

\begin{equation}
\label{eqn:h2}
    SF_{H_{2}} = (w\_s) F_{H_{1}} + (1-w\_s) F_{H_{2}}
\end{equation}

From H3 onwards, either partial or full interpolation can be performed to stabilise the forecasts. Equations \ref{eqn:hpartial} and \ref{eqn:hfull} show the formulas of partial and full interpolation used to make the forecasts horizontally stable, where $SF_{H_{j}}$ is the stable forecast of the $j^{th}$ horizon, $F_{H_{j}}$ is the original forecast of the $j^{th}$ horizon and $w\_s$ is the weight given for the forecast of the $(j-1)^{th}$ horizon, where $2 < j \leq h$ and $0 \leq w\_s \leq 1$. 

\begin{equation}
\label{eqn:hpartial}
    SF_{H_{j}} = (w\_s) F_{H_{j-1}} + (1-w\_s) F_{H_{j}}
\end{equation}

\begin{equation}
\label{eqn:hfull}
    SF_{H_{j}} = (w\_s) SF_{H_{j-1}} + (1-w\_s) F_{H_{j}}
\end{equation}

Similar to vertical stability, here also the difference between partial and full interpolation methods is that to obtain stable forecasts for the current horizon, partial interpolation considers the corresponding original forecasts from the previous horizon whereas full interpolation considers the interpolated stable forecasts from the previous horizon. 

\subsubsection{Linear Interpolation for Joint Vertical and Horizontal Stability}

The linear interpolation approach is directly applicable to stabilise forecasts both vertically and horizontally at the same time by interpolating between the respective forecasts. As such, we can either directly interpolate between the respective forecasts, which are the diagonals shown in the bottom-right quadrant of Figure~\ref{fig:stable_intro}, or we can first stabilise vertically then horizontally, or vice versa. Doing two separate steps has the advantage of using two different interpolation weights and is able to control for the degree of vertical and horizontal stabilisation separately.

\bigskip
For our experiments, we focus on either horizontal or vertical stability separately and analyse the effect of both partial and full interpolation methods to gain vertical and horizontal stability with different weights for $w\_s$, namely 0.2, 0.4, 0.5, 0.6, 0.8, and 1.

\section{Experimental Framework}
\label{sec:experiments}
In this section, we discuss the datasets, error measures, base models, and benchmarks used in our experiments. 

\subsection{Datasets}
\label{sec:datasets}
We use four publicly available datasets\footnote{The experimental datasets are available at \url{https://github.com/rakshitha123/StableForecasting}.} to evaluate the performance of our proposed framework. Table \ref{tab:dataset_overview} provides a summary of the datasets, a brief overview is as follows.

\begin{itemize}
    \item M4 Monthly Dataset: The monthly dataset of the M4 forecasting competition \citep{makridakis_2018_m4}.
    \item M3 Monthly Dataset: The monthly dataset of the M3 forecasting competition \citep{makridakis_2000_m3}.
    \item Favorita Dataset \citep{favourita_2018_kaggle}:  The first 1000 time series from the Corporaci\'on Favorita Grocery Sales forecasting competition. Each series shows the daily unit sales of a particular item sold at a Favorita store. Missing observations in this dataset are replaced by zeros. 
    \item M5 Items Dataset: An aggregated version of the M5 forecasting competition dataset \citep{makridakis_2020_m5} where the daily unit sales of individual items have been aggregated across Walmart stores in different states.
\end{itemize}
   
\begin{table*}
\renewcommand{\arraystretch}{0.6}
\centering
\begin{tabular}{lccccc}
\hline
\textbf{Dataset} & \textbf{No. of} & \textbf{No. of} & \textbf{Forecast} &  \textbf{Min.} & \textbf{Max.} \\
 & \textbf{Series} & \textbf{Origins} & \textbf{Horizon} &  \textbf{Length} & \textbf{Length} \\

\hline
M4 &  48000 & 13 & 6 & 60 & 2812 \\
M3 &  1428 & 13 & 6 & 66 & 144 \\
Favorita &  1000 &  11  & 6  & 1684  &  1684  \\
M5 & 3049  &  13  & 16  & 1969  &  1969  \\
\hline
\end{tabular}
\caption{Summary of the used datasets.}
\label{tab:dataset_overview}
\end{table*}

The major reason for using the M3 and M4 monthly datasets is to compare the performance of the proposed approach against the Stable N-BEATS approach \citep{VANBELLE20231333}. Those authors have used all the sub-datasets of the M3 and M4 forecasting competitions for their experiments. For a meaningful comparison, we have used the monthly versions of both datasets which are the corresponding largest sub-datasets in terms of the number of time series. Furthermore, to add some diversity to the pool of datasets, we consider the Favorita and M5 items datasets, which are real-world retail datasets where our proposed approach is highly useful in practice.

\subsection{Error Measures}
\label{sec:error_measure}

We use different error measures to evaluate forecast accuracy and stability. These error measures are explained in the following.

\subsubsection{Accuracy Measures}
\label{sec:acc_measures}
The accuracy of the forecasts is evaluated using error measures that are common in the forecasting research space: Mean Absolute Scaled Error \citep[MASE, ][]{ref_36} and Root Mean Squared Scaled Error \citep[RMSSE, ][]{makridakis_2020_m5}, which are defined in Equations \ref{eqn:mase} and \ref{eqn:rmsse}, respectively. For a given dataset, forecast accuracy is evaluated across series with multiple forecast origins. Thus,  the error measures are defined across one to $h$-step ahead forecasts resulting from a specific forecasting origin $t$, where $h$ is the forecast horizon, $m$ is the seasonal period of the dataset, $y_{t+i}$ is the actual series value at time $t+i$ and $\hat{y}_{t+i|t}$ is the forecast corresponding with time $t+i$ made at time $t$.

\begin{equation}
\label{eqn:mase}
    \textit{MASE} = \frac{\sum_{i=1}^{h} |y_{t+i} - \hat{y}_{t+i|t}|}{\frac{h}{t - m}\sum_{i=m+1}^{t} |y_{i} - y_{i-m}|} 
\end{equation}

\begin{equation}
\label{eqn:rmsse}
    \textit{RMSSE} = \sqrt{\frac{\sum_{i=1}^{h} (y_{t+i} - \hat{y}_{t+i|t})^2}{\frac{h}{t - m}\sum_{i=m+1}^{t} (y_{i} - y_{i-m})^2}} 
\end{equation}

Note that we use MASE as the measure for the M3 and M4 monthly datasets, and the RMSSE for the M5 and Favorita datasets. This is in line with standard practices in the literature. 
A version of RMSSE was used in the M5 competition due to its intermittency, whereas MASE was used in the M4 \citep[e.g.,][]{makridakis_2020_m5}. For intermittent data, the MASE would favour all-zero forecasts, whereas the RMSSE penalises larger errors more due to its quadratic nature. From the datasets used in our evaluation, the Favorita dataset is intermittent, and although we use an aggregated version of the M5, the series are still intermittent here as well.

To measure the performance of the models on a dataset, we calculate the mean values of MASE and RMSSE across multiple forecast origins in all series. Thus, the accuracy of each model is finally evaluated using the mean MASE and mean RMSSE across a dataset.

\subsubsection{Stability Measures}
\label{sec:stab_measures}
The stability of the forecasts is evaluated using different error measures that follow the stability measures introduced by \citet{VANBELLE20231333}. The definitions of the error measures are changed for the vertical and horizontal stability types, which are explained in the following. 
To measure vertical stability, \citet{VANBELLE20231333} propose the symmetric Mean Absolute Percentage Change (sMAPC), which is based on sMAPE and is defined in Equation \ref{eqn:smapc}. 

\begin{equation}
\label{eqn:smapc}
    \textit{sMAPC}(V) = \frac{200\%}{(h-1)}\sum_{i=1}^{h-1} \frac{|\hat{y}_{t+i|t} - \hat{y}_{t+i|t-1}|}{|\hat{y}_{t+i|t}| + |\hat{y}_{t+i|t-1}|} 
\end{equation}

Here, sMAPC measures the change in one-to-$h$-step-ahead forecasts corresponding to two adjacent forecast origins, $t$ and $t-1$. Thus, it provides a measurement of the extent to which the forecasts generated at origin $t$ are stable compared with the forecasts generated at origin $t-1$ for the same time periods. In line with the definition of sMAPC, we define the two stability measures used in our work, Mean Absolute Scaled Change (MASC) and Root Mean Squared Scaled Change (RMSSC), which compare the change between the forecasts generated at two adjacent forecast origins for the same time periods. MASC and RMSSC are defined in Equations \ref{eqn:masc} and \ref{eqn:rmssc}, respectively.

\begin{equation}
\label{eqn:masc}
    \textit{MASC}(V) = \frac{\sum_{i=1}^{h-1} |\hat{y}_{t+i|t} - \hat{y}_{t+i|t-1}|}{\frac{h-1}{t-m- 1}\sum_{i=m+1}^{t-1} |y_{i} - y_{i-m}|} 
\end{equation}

\begin{equation}
\label{eqn:rmssc}
    \textit{RMSSC}(V) = \sqrt{\frac{\sum_{i=1}^{h-1} (\hat{y}_{t+i|t} - \hat{y}_{t+i|t-1})^2}{\frac{h-1}{t-m-1}\sum_{i=m+1}^{t-1} (y_{i} - y_{i-m})^2}} 
\end{equation}

We also measure vertical forecast stability in terms of the change between the forecasts generated at origin $t$ and the first set of forecasts generated for the same time periods at a previous origin. We refer to MASC and RMSSC calculated in this way as MASC\_I and RMSSC\_I, respectively. When calculating MASC\_I and RMSSC\_I for vertical stability, the term $\hat{y}_{t+i|t-1}$ in Equations \ref{eqn:masc} and \ref{eqn:rmssc} is replaced with the first forecast generated for time $t+i$ at a previous origin. 

Horizontal stability measures the change in the forecasts generated at the same origin. Thus, to measure the horizontal forecast stability in the most generic case of stabilising towards a flat line, the definitions of MASC and RMSSC are respectively changed as shown in Equations \ref{eqn:masch} and \ref{eqn:rmssch}.

\begin{equation}
\label{eqn:masch}
    \textit{MASC}(H) = \frac{\sum_{i=2}^{h} |\hat{y}_{t+i|t} - \hat{y}_{t+i-1|t}|}{\frac{h-1}{t-m}\sum_{i=m+1}^{t} |y_{i} - y_{i-m}|} 
\end{equation}

\begin{equation}
\label{eqn:rmssch}
    \textit{RMSSC}(H) = \sqrt{\frac{\sum_{i=2}^{h} (\hat{y}_{t+i|t} - \hat{y}_{t+i-1|t})^2}{\frac{h-1}{t-m}\sum_{i=m+1}^{t} (y_{i} - y_{i-m})^2}} 
\end{equation}

The MASC\_I and RMSSC\_I are also redefined for horizontal stability. For a given origin $t$, these stability measures calculate the change between the forecasts for time $t+2$ onwards and the first forecast at time $t+1$. Thus, to calculate MASC\_I and RMSSC\_I for horizontal stability, the term $\hat{y}_{t+i-1|t}$ in Equations \ref{eqn:masch} and \ref{eqn:rmssch} is replaced with $\hat{y}_{t+1|t}$. 

For a given dataset, all stability measures are calculated for each series and origin. Thus, to measure the forecast stability of the models on a dataset, the mean values of MASC, RMSSC, MASC\_I, and RMSSC\_I are calculated across multiple forecast origins in all the series. Thus, the vertical and horizontal stabilities of each model are evaluated using the following error measures across a dataset: mean MASC, mean RMSSC, mean MASC\_I, and mean RMSSC\_I. In the remainder of the paper, the names of the error measures are not accompanied by the term \textit{mean}.

\subsection{Experimental Base Models and Benchmarks}
\label{sec:base_models}

Our proposed framework is model-agnostic (with some limitations in the horizontal case, as discussed in Section~\ref{sec:hor_int}) and it is applicable to stabilise the forecasts obtained from any forecasting model; in particular, it is applicable to local models, global models, and even judgemental forecasts that are not based on data. For the experiments, we use the following base models: N-BEATS \citep{oreshkin_2019_nbeats}, Pooled Regression \citep[PR,][]{gelman_hill_2006, pablo_2020_principles} and LightGBM \citep{guolin_lightgbm_2017} as global models, and ARIMA and ETS as local models \citep{ref_4}. The method proposed by \citet{VANBELLE20231333} is applicable to stabilise the forecasts of N-BEATS and thus, we use N-BEATS as a base model for the experiments. LightGBM is a highly efficient gradient-boosted tree that has recently gained popularity in the forecasting domain after contributing to most of the top solutions of the M5 forecasting competition \citep{makridakis_2020_m5}. To further add some diversity to the pool of base models, we consider PR as a base model, which is a globally trained linear model, and with ETS and ARIMA, some well-established standard local models.

We use the N-BEATS implementation by \citet{VANBELLE20231333} for the experiments. The original N-BEATS model is executed by setting the hyperparameter $\lambda$ in the implementation to zero. The Stable N-BEATS model is also executed as a benchmark by setting $\lambda$ to the corresponding optimal values provided by \citet{VANBELLE20231333}. The remaining hyperparameters of the N-BEATS model are also set to those given by \citet{VANBELLE20231333}. 
As the performance of the N-BEATS implementation depends to a certain degree on the hyperparameters, the N-BEATS models are only executed across the M3 and M4 monthly datasets, where the optimal values of all hyperparameters are available in \citet{VANBELLE20231333}. 

We use the automatic versions of ETS and ARIMA in their implementations in the \verb|forecast| package in R~\citep{ref_22} with their default parameter settings.
The R packages \verb|glmnet| \citep{ref_111} and \verb|lightgbm| \citep{ke_2020_lightgbm} are used to implement the PR and LightGBM models. The LightGBM model is executed with the default hyperparameter values except for the learning rate, minimum instances in a leaf node, and the number of estimators, where the values of these parameters are respectively set to 0.075, 100, and 100. The PR model does not require hyperparameters. The LightGBM and PR models are executed across all datasets, using only lagged values as features/inputs. 
To perform multi-step-ahead forecasting, we use an iterated one-step-ahead approach where forecasts are fed back iteratively as inputs to the algorithm, for the lags that are not yet available.
The number of lagged values used in PR and LightGBM are determined using a heuristic suggested by \citet{ref_6}. In particular, the number of lags is set to be the $\text{seasonal period of the dataset} \times 1.25$. Thus, we consider 9 and 15 lags for the daily and monthly datasets, respectively.

The original N-BEATS, Stable N-BEATS, PR, LightGBM, ETS, and ARIMA models are also considered as the main benchmarks of this study.

\subsection{Statistical Tests of the Results}
\label{sec:statistical_testing}
We also perform pairwise tests for statistical significance using a Wilcoxon test \citep{rey_2011_stats}, with an initial significance level of $\alpha = 0.05$. A Bonferroni correction is applied for $\alpha$ by dividing it by the number of comparisons made (for details, see Section \ref{sec:stat_results}).

\section{Results and Discussion}
\label{sec:eval_accuracy}

This section explains the results of the vertical and horizontal stability experiments and provides more insights regarding the proposed models. The accuracy results are reported in terms of MASE (for M3 and M4 datasets) and RMSSE (for the intermittent Favorita and M5 datasets), and stability results are reported in terms of MASC, RMSSC, MASC\_I, and RMSSC\_I. 
The Online Appendix\footnote{The online appendix is available at \url{https://github.com/rakshitha123/StableForecasting}.} contains the results for all datasets for MASE, RMSSE, and sMAPE as well as the corresponding stability measures.

In the results tables, the terms PI and FI denote partial and full interpolation experiments, respectively. The numerical values next to the terms PI and FI show the corresponding w\_s values considered during interpolation.

\subsection{Results of Vertical Stability Experiments}

Table \ref{tab:ver_results} shows the results of the vertical stability experiments for all the experimental datasets. For the M4 and M3 datasets, the results are reported for MASE, MASC and MASC\_I, whereas for the Favorita and M5 datasets, the results are reported for RMSSE, RMSSC and RMSSC\_I.  The experiments with the five base models, N-BEATS, PR, LightGBM, ETS, and ARIMA are separately grouped. The sub-experiments related to each base model are further divided into three groups. The first sub-group contains the benchmarks, the base models that do not use any stabilisation techniques. For the N-BEATS models, the first sub-group also contains the Stable N-BEATS approach. The second and third sub-groups, respectively, show the partial and full interpolation experiments performed with the corresponding base model forecasts. The results of the best-performing variants in each sub-group (Base, PI, FI) are highlighted in boldface italicised font, and the overall best-performing variants corresponding to a particular base model across all sub-groups are highlighted in boldface only. Across the PR, LightGBM, ETS, and ARIMA results, the values of the best model in the base models sub-group are not shown in boldface italicised as those sub-groups only contain one model in each, so that this highlighting would not be informative. Thus, in those sub-groups, a value is only highlighted in boldface if the corresponding base model is the best-performing model across all groups. 

\begin{table}
        \vspace{-9.5em}
        \caption{Results of vertical stability experiments across all datasets.  The * symbols represent the results that are significantly more accurate and stable than the corresponding base model results. Boldface only shows the best model across all sub-groups (PI, FI, Base) of a particular base model, whereas boldface italics show the best model within a particular sub-group (where the Base sub-group only contains one model the highlighting for this case is omitted).
        }
        
\begin{center}
        \resizebox{\textwidth}{!}{
		\begin{tabular}{lrrrrrrrrrrrrr}
			\toprule
			& & \multicolumn{4}{c}{MASE/RMSSE} & \multicolumn{4}{c}{MASC/RMSSC} & \multicolumn{4}{c}{MASC\_I/RMSSC\_I} \\
			\cmidrule(lr){3-6} \cmidrule(lr){7-10} \cmidrule(lr){11-14}
			& & M4 & M3 & Favorita  & M5 & M4 & M3 & Favorita  & M5 & M4 & M3 & Favorita  & M5  \\
      \hline
			 \addlinespace

N-BEATS & Base & \textbf{\textit{0.638}} & 0.643 & - & - & 0.307 & 0.260 & - & - & 0.456 & 0.392 & - & - \\ 
   & Stable & 0.648 & $\textbf{0.639}^*$ & - & - & \textbf{\textit{0.253}} & $\textbf{\textit{0.193}}^*$ & - & - & \textbf{\textit{0.416}} & $\textbf{\textit{0.331}}^*$ & - & - \\ 
   \cmidrule(lr){2-14}
   & PI\_0.2 & \textbf{\textit{0.635}} & \textbf{\textit{0.642}} & - & - & 0.276 & 0.211 & - & - & 0.436 & 0.359 & - & - \\ 
   & PI\_0.4 & 0.638 & 0.645 & - & - & 0.224 & 0.177 & - & - & 0.398 & 0.330 & - & - \\ 
   & PI\_0.5 & 0.643 & 0.648 & - & - & 0.210 & 0.168 & - & - & 0.382 & 0.317 & - & - \\ 
   & PI\_0.6 & 0.649 & 0.652 & - & - & \textbf{\textit{0.206}} & \textbf{\textit{0.164}} & - & - & 0.368 & 0.305 & - & - \\ 
   & PI\_0.8 & 0.665 & 0.662 & - & - & 0.220 & 0.172 & - & - & 0.345 & 0.285 & - & - \\ 
   & PI\_1 & 0.687 & 0.675 & - & - & 0.255 & 0.192 & - & - & \textbf{\textit{0.330}} & \textbf{\textit{0.271}} & - & - \\ 
   \cmidrule(lr){2-14}
   & FI\_0.2 & $\textbf{0.634}^*$ & \textbf{\textit{0.642}} & - & - & $0.275^*$ & 0.210 & - & - & $0.430^*$ & 0.354 & - & - \\ 
   & FI\_0.4 & 0.637 & 0.644 & - & - & 0.210 & 0.163 & - & - & 0.369 & 0.305 & - & - \\ 
   & FI\_0.5 & 0.642 & 0.647 & - & - & 0.178 & 0.140 & - & - & 0.332 & 0.274 & - & - \\ 
   & FI\_0.6 & 0.651 & 0.651 & - & - & 0.147 & 0.117 & - & - & 0.288 & 0.238 & - & - \\ 
   & FI\_0.8 & 0.683 & 0.670 & - & - & 0.082 & 0.066 & - & - & 0.171 & 0.142 & - & - \\ 
   & FI\_1 & 0.753 & 0.716 & - & - & \textbf{0.000} & \textbf{0.000} & - & - & \textbf{0.000} & \textbf{0.000} & - & - \\ 
            \Xhline{5\arrayrulewidth}
			 \addlinespace

PR & Base & \textbf{0.791} & \textbf{0.755} & 0.586 & 1.009 & 0.219 & 0.190 & 0.106 & 0.251 & 0.372 & 0.287 & 0.133 & 0.437 \\ 
\cmidrule(lr){2-14}
   & PI\_0.2 & \textbf{\textit{0.798}} & \textbf{\textit{0.759}} & 0.585 & 1.003 & 0.197 & 0.157 & 0.084 & 0.198 & 0.355 & 0.260 & 0.119 & 0.412 \\ 
   & PI\_0.4 & 0.808 & 0.765 & $\textbf{0.584}^*$ & \textbf{\textit{1.002}} & 0.170 & 0.132 & $0.066^*$ & 0.161 & 0.323 & 0.234 & $0.107^*$ & 0.392 \\ 
   & PI\_0.5 & 0.813 & 0.768 & $\textbf{0.584}^*$ & \textbf{\textit{1.002}} & 0.160 & 0.124 & $0.061^*$ & 0.153 & 0.308 & 0.223 & $0.102^*$ & 0.384 \\ 
   & PI\_0.6 & 0.819 & 0.772 & 0.585 & 1.003 & 0.155 & \textbf{\textit{0.119}} & \textbf{\textit{0.058}} & \textbf{\textit{0.152}} & 0.295 & 0.212 & 0.097 & 0.378 \\ 
   & PI\_0.8 & 0.832 & 0.781 & 0.586 & 1.007 & \textbf{\textit{0.152}} & \textbf{\textit{0.119}} & 0.059 & 0.170 & 0.270 & 0.194 & 0.090 & \textbf{\textit{0.371}} \\ 
   & PI\_1 & 0.847 & 0.791 & 0.589 & 1.015 & 0.160 & 0.129 & 0.070 & 0.211 & \textbf{\textit{0.250}} & \textbf{\textit{0.180}} & \textbf{\textit{0.088}} & 0.372 \\ 
   \cmidrule(lr){2-14}
   & FI\_0.2 & \textbf{\textit{0.799}} & \textbf{\textit{0.760}} & 0.585 & 1.003 & 0.195 & 0.156 & 0.084 & 0.197 & 0.349 & 0.256 & 0.118 & 0.408 \\ 
   & FI\_0.4 & 0.811 & 0.768 & $\textbf{0.584}^*$ & 0.999 & 0.157 & 0.123 & $0.063^*$ & 0.150 & 0.299 & 0.217 & $0.100^*$ & 0.373 \\ 
   & FI\_0.5 & 0.819 & 0.773 & $\textbf{0.584}^*$ & 0.997 & 0.137 & 0.106 & $0.053^*$ & 0.128 & 0.268 & 0.193 & $0.089^*$ & 0.351 \\ 
   & FI\_0.6 & 0.829 & 0.780 & 0.585 & $\textbf{0.996}^*$ & 0.115 & 0.089 & 0.043 & $0.107^*$ & 0.231 & 0.166 & 0.077 & $0.322^*$ \\ 
   & FI\_0.8 & 0.859 & 0.800 & 0.587 & $\textbf{0.996}^*$ & 0.066 & 0.050 & 0.023 & $0.062^*$ & 0.136 & 0.096 & 0.045 & $0.227^*$ \\ 
   & FI\_1 & 0.911 & 0.833 & 0.594 & 1.027 & \textbf{0.000} & \textbf{0.000} & \textbf{0.000} & \textbf{0.000} & \textbf{0.000} & \textbf{0.000} & \textbf{0.000} & \textbf{0.000} \\ 
			\Xhline{5\arrayrulewidth}
			 \addlinespace

LightGBM & Base & \textbf{0.854} & \textbf{0.766} & \textbf{0.631} & 0.987 & 0.281 & 0.240 & 0.095 & 0.180 & 0.445 & 0.350 & 0.146 & 0.355 \\ 
\cmidrule(lr){2-14}
   & PI\_0.2 & \textbf{\textit{0.858}} & \textbf{\textit{0.769}} & \textbf{\textit{0.632}} & \textbf{\textit{0.985}} & 0.261 & 0.194 & 0.078 & 0.144 & 0.425 & 0.318 & 0.133 & 0.337 \\ 
   & PI\_0.4 & 0.866 & 0.775 & 0.634 & \textbf{\textit{0.985}} & 0.215 & 0.160 & 0.064 & 0.121 & 0.387 & 0.289 & 0.121 & 0.323 \\ 
   & PI\_0.5 & 0.872 & 0.778 & 0.635 & 0.986 & 0.202 & 0.150 & 0.060 & \textbf{\textit{0.116}} & 0.370 & 0.277 & 0.115 & 0.317 \\ 
   & PI\_0.6 & 0.879 & 0.783 & 0.636 & 0.987 & \textbf{\textit{0.196}} & \textbf{\textit{0.146}} & \textbf{\textit{0.058}} & \textbf{\textit{0.116}} & 0.354 & 0.265 & 0.110 & 0.312 \\ 
   & PI\_0.8 & 0.896 & 0.793 & 0.639 & 0.990 & 0.202 & 0.151 & \textbf{\textit{0.058}} & 0.128 & 0.327 & 0.245 & 0.102 & 0.305 \\ 
   & PI\_1 & 0.917 & 0.806 & 0.642 & 0.995 & 0.226 & 0.170 & 0.066 & 0.154 & \textbf{\textit{0.308}} & \textbf{\textit{0.231}} & \textbf{\textit{0.097}} & \textbf{\textit{0.303}} \\ 
   \cmidrule(lr){2-14}
   & FI\_0.2 & \textbf{\textit{0.859}} & \textbf{\textit{0.770}} & \textbf{\textit{0.633}} & 0.985 & 0.259 & 0.193 & 0.077 & 0.143 & 0.420 & 0.314 & 0.131 & 0.334 \\ 
   & FI\_0.4 & 0.869 & 0.777 & 0.635 & $\textbf{0.984}^*$ & 0.202 & 0.150 & 0.061 & $0.112^*$ & 0.358 & 0.268 & 0.112 & $0.307^*$ \\ 
   & FI\_0.5 & 0.878 & 0.782 & 0.637 & $\textbf{0.984}^*$ & 0.174 & 0.129 & 0.052 & $0.097^*$ & 0.321 & 0.240 & 0.101 & $0.290^*$ \\ 
   & FI\_0.6 & 0.889 & 0.790 & 0.639 & $\textbf{0.984}^*$ & 0.144 & 0.107 & 0.044 & $0.081^*$ & 0.277 & 0.207 & 0.087 & $0.266^*$ \\ 
   & FI\_0.8 & 0.924 & 0.813 & 0.645 & 0.986 & 0.080 & 0.060 & 0.025 & 0.049 & 0.164 & 0.122 & 0.052 & 0.188 \\ 
   & FI\_1 & 0.990 & 0.857 & 0.657 & 1.014 & \textbf{0.000} & \textbf{0.000} & \textbf{0.000} & \textbf{0.000} & \textbf{0.000} & \textbf{0.000} & \textbf{0.000} & \textbf{0.000} \\ 
            \Xhline{5\arrayrulewidth}
			 \addlinespace

 ETS & Base & \textbf{0.656} & \textbf{0.616} & 0.567 & 0.933 & 0.263 & 0.209 & 0.053 & 0.102 & 0.453 & 0.337 & 0.090 & 0.212 \\ 
 \cmidrule(lr){2-14}
   & PI\_0.2 & \textbf{\textit{0.657}} & \textbf{\textit{0.617}} & \textbf{\textit{0.567}} & \textbf{\textit{0.933}} & 0.242 & 0.171 & 0.043 & 0.084 & 0.440 & 0.311 & 0.084 & 0.202 \\ 
   & PI\_0.4 & 0.664 & 0.621 & \textbf{\textit{0.567}} & 0.934 & 0.207 & 0.145 & 0.037 & 0.073 & 0.405 & 0.287 & 0.078 & 0.194 \\ 
   & PI\_0.5 & 0.670 & 0.624 & 0.568 & 0.934 & 0.197 & 0.138 & 0.036 & 0.070 & 0.389 & 0.276 & 0.075 & 0.190 \\ 
   & PI\_0.6 & 0.676 & 0.628 & 0.568 & 0.935 & \textbf{\textit{0.193}} & \textbf{\textit{0.136}} & \textbf{\textit{0.035}} & \textbf{\textit{0.069}} & 0.374 & 0.265 & 0.073 & 0.187 \\ 
   & PI\_0.8 & 0.693 & 0.638 & 0.569 & 0.937 & 0.199 & 0.143 & 0.037 & 0.076 & 0.349 & 0.248 & 0.070 & 0.183 \\ 
   & PI\_1 & 0.713 & 0.651 & 0.570 & 0.940 & 0.218 & 0.158 & 0.043 & 0.089 & \textbf{\textit{0.329}} & \textbf{\textit{0.234}} & \textbf{\textit{0.068}} & \textbf{\textit{0.181}} \\ 
   \cmidrule(lr){2-14}
   & FI\_0.2 & \textbf{\textit{0.657}} & \textbf{\textit{0.617}} & $\textbf{0.566}^*$ & $\textbf{0.932}^*$ & 0.240 & 0.170 & $0.043^*$ & $0.083^*$ & 0.433 & 0.306 & $0.082^*$ & $0.200^*$ \\ 
   & FI\_0.4 & 0.663 & 0.621 & 0.567 & 0.933 & 0.190 & 0.133 & 0.034 & 0.066 & 0.375 & 0.266 & 0.072 & 0.184 \\ 
   & FI\_0.5 & 0.670 & 0.625 & 0.567 & 0.934 & 0.164 & 0.116 & 0.030 & 0.058 & 0.338 & 0.240 & 0.066 & 0.173 \\ 
   & FI\_0.6 & 0.679 & 0.631 & 0.568 & 0.934 & 0.138 & 0.097 & 0.025 & 0.049 & 0.293 & 0.209 & 0.057 & 0.158 \\ 
   & FI\_0.8 & 0.714 & 0.652 & 0.569 & 0.937 & 0.079 & 0.056 & 0.015 & 0.029 & 0.175 & 0.125 & 0.035 & 0.111 \\ 
   & FI\_1 & 0.790 & 0.700 & 0.574 & 0.952 & \textbf{0.000} & \textbf{0.000} & \textbf{0.000} & \textbf{0.000} & \textbf{0.000} & \textbf{0.000} & \textbf{0.000} & \textbf{0.000} \\
            \Xhline{5\arrayrulewidth}
			 \addlinespace

ARIMA & Base & \textbf{0.634} & \textbf{0.618} & 0.573 & 0.951 & 0.275 & 0.206 & 0.064 & 0.096 & 0.435 & 0.326 & 0.085 & 0.203 \\ 
\cmidrule(lr){2-14}
   & PI\_0.2 & \textbf{\textit{0.635}} & \textbf{\textit{0.619}} & \textbf{\textit{0.572}} & \textbf{\textit{0.951}} & 0.225 & 0.168 & 0.051 & 0.078 & 0.398 & 0.299 & 0.077 & 0.192 \\ 
   & PI\_0.4 & 0.641 & 0.622 & \textbf{\textit{0.572}} & \textbf{\textit{0.951}} & 0.191 & 0.142 & 0.042 & 0.067 & 0.365 & 0.275 & 0.069 & 0.184 \\ 
   & PI\_0.5 & 0.646 & 0.625 & \textbf{\textit{0.572}} & 0.952 & 0.181 & 0.135 & 0.038 & 0.064 & 0.350 & 0.264 & 0.066 & 0.180 \\ 
   & PI\_0.6 & 0.653 & 0.629 & \textbf{\textit{0.572}} & 0.953 & \textbf{\textit{0.177}} & \textbf{\textit{0.132}} & \textbf{\textit{0.037}} & \textbf{\textit{0.063}} & 0.336 & 0.253 & 0.063 & 0.177 \\ 
   & PI\_0.8 & 0.668 & 0.639 & 0.573 & 0.954 & 0.181 & 0.137 & 0.038 & 0.067 & 0.312 & 0.235 & 0.059 & 0.172 \\ 
   & PI\_1 & 0.688 & 0.651 & 0.575 & 0.957 & 0.200 & 0.151 & 0.044 & 0.079 & \textbf{\textit{0.294}} & \textbf{\textit{0.222}} & \textbf{\textit{0.058}} & \textbf{\textit{0.170}} \\ 
   \cmidrule(lr){2-14}
   & FI\_0.2 & \textbf{\textit{0.635}} & \textbf{\textit{0.619}} & 0.572 & $\textbf{0.950}^*$ & 0.223 & 0.167 & 0.051 & $0.078^*$ & 0.393 & 0.295 & 0.076 & $0.191^*$ \\ 
   & FI\_0.4 & 0.642 & 0.622 & $\textbf{0.571}^*$ & 0.951 & 0.176 & 0.131 & $0.039^*$ & 0.061 & 0.338 & 0.255 & $0.064^*$ & 0.175 \\ 
   & FI\_0.5 & 0.648 & 0.626 & 0.572 & 0.952 & 0.152 & 0.113 & 0.033 & 0.054 & 0.304 & 0.229 & 0.057 & 0.165 \\ 
   & FI\_0.6 & 0.658 & 0.632 & 0.572 & 0.952 & 0.127 & 0.095 & 0.027 & 0.046 & 0.264 & 0.199 & 0.049 & 0.151 \\ 
   & FI\_0.8 & 0.692 & 0.652 & 0.573 & 0.956 & 0.073 & 0.054 & 0.015 & 0.028 & 0.157 & 0.119 & 0.029 & 0.106 \\ 
   & FI\_1 & 0.761 & 0.698 & 0.577 & 0.970 & \textbf{0.000} & \textbf{0.000} & \textbf{0.000} & \textbf{0.000} & \textbf{0.000} & \textbf{0.000} & \textbf{0.000} & \textbf{0.000} \\
\Xhline{5\arrayrulewidth}
			\end{tabular}
		}	
		\label{tab:ver_results}
\end{center}
\end{table}

In the N-BEATS model group, our proposed model with settings FI\_0.2 and PI\_0.2 achieves better performance in terms of both accuracy and stability compared to the base N-BEATS model which does not use any stabilisation techniques, across both the M3 and M4 datasets. Thus, here we can achieve more stable models that are also more accurate and do not need to trade stability for a lower accuracy.
The Stable N-BEATS and FI\_0.2 variant show the best performance across the M3 and M4 datasets, respectively, in terms of MASE. Thus, in terms of accuracy, our proposed framework and Stable N-BEATS approach share the lead. In terms of MASC and MASC\_I, the Stable N-BEATS model is always better than FI\_0.2 across both the M3 and M4 datasets. However, using our proposed methods, the trade-off between stability and accuracy can be controlled easily, and it is possible to make the forecasts stable to any extent as required using the value of w\_s. For example, the FI\_1 method always provides completely stable forecasts with any base model as it results in the forecasts made at the current origin being the same as the forecasts made at the corresponding previous origin by setting the value of w\_s to 1. Thus, with our framework, practitioners can choose whether they need more accurate forecasts or more stable forecasts based on the requirement, and choose w\_s accordingly, without having to re-run the base forecasting models. With the Stable N-BEATS implementation \citep{VANBELLE20231333} also it would be possible to make forecasts more stable with the value of $\lambda$. 
However, the original paper tuned $\lambda$ as a hyperparameter to increase stability while minimising the loss of accuracy, and the method needs to be retrained when $\lambda$ is changed. 
Furthermore, the method does not allow direct control of stability in terms of how close new forecasts should be to already communicated forecasts from prior origins. In particular, forecasts cannot be made arbitrarily stable in a meaningful way, as choosing very high values for $\lambda$ will lead to forecasts that are always zero (which makes them perfectly stable, but not accurate).

In the PR and LightGBM model groups, the base models and our proposed models show a mixed performance in terms of accuracy, where the base models show better performance across the M3 and M4 datasets, as well  as Favorita for LightGBM, and our proposed models show better performance across the M5 dataset, and Favorita for PR. However, our proposed models are always considerably more stable than the base models across all stability measures, with modest accuracy losses.

In terms of stability, the full interpolation models are better overall than the partial interpolation models across all datasets for all stability measures. The full interpolation models use the interpolated stable forecasts of the previous origin to produce forecasts for the current origin and thus provide more stable forecasts compared to the partial interpolation models. 

However, the accuracy of the partial and full interpolation experiments varies with no clear winner. For all datasets and methods, full interpolation models usually provide slightly better accuracy than partial interpolation models for lower degrees of stability, and accuracy degrades when increasing the degree of stability.

Variants with higher w\_s provide more stable forecasts in terms of MASC/RMSSC and MASC\_I/RMSSC\_I across all datasets with all base models. When w\_s is high, the variants consider a higher proportion of the forecasts made at the previous origin to produce forecasts at the current origin, which is the main reason for this phenomenon. However, in terms of accuracy, variants with different w\_s values show the best performance across different datasets. The best partial and full interpolation variants across the M3 and M4 datasets in terms of accuracy use 0.2 for w\_s, which is the lowest weight considered in the experiments. The lowest w\_s value does not make the forecasts considerably more stable as it only considers a small proportion of the forecasts made at the previous origin to make the forecasts at the current origin. Thus, out of the proposed model variants, PI\_0.2 and FI\_0.2 overall produce the least stable forecasts. However, these variants produce the most accurate forecasts for the M3 and M4 datasets. In comparison, the most accurate partial and full interpolation variants across the Favorita and M5 datasets tend to use higher values for w\_s. In particular, the most accurate model across the M5 dataset with the PR and LightGBM is FI\_0.8 and FI\_0.6, respectively. Higher w\_s values provide more stable forecasts. We hypothesise that more stable forecasts are more accurate for the Favorita and M5 datasets, as these are intermittent datasets and thus, their series tend to have lower variance compared with typical non-intermittent series, as the series consist mainly of zeros and small positive integer values. In particular, the mean per series standard deviations are below 10 for the Favorita and M5 datasets and above 1000 for the versions of M3 and M4 that we use. Making forecasts vertically stable also reduces the variance of forecasts as it aims to bring the forecasts closer to the corresponding forecasts generated at the previous origin. Thus, the stabilised forecasts are more accurate for those datasets with the considered base models. In comparison, the M3 and M4 datasets have higher degrees of trends and seasonality \citep{godahewa_2021_monash}. Making forecasts vertically stable can reduce the trends and seasonality of the forecasts; thus, stable forecasts tend to not be the most accurate forecasts for these datasets.

Finally, regarding the question of how stability interacts with the forecast horizon, the Online Appendix B\footnote{The online appendix is available at \url{https://github.com/rakshitha123/StableForecasting}} shows some plots. There are situations when with increasing horizon the forecasts become more stable, whereas in other cases they become less stable, and there is no clear picture, in contrast to accuracy, which always decreases with increasing horizon.

\begin{table}
        \vspace{-9.5em}
        \caption{Results of horizontal stability experiments across all datasets.  The * symbols represent the results that are significantly more accurate and stable than the corresponding base model results. Boldface only shows the best model across all sub-groups (PI, FI, Base) of a particular base model, whereas boldface italics show the best model within a particular sub-group (where the Base sub-group only contains one model the highlighting is omitted).
        }
        
\begin{center}
        \resizebox{\textwidth}{!}{
		\begin{tabular}{lrrrrrrrrrrrrr}
			\toprule
			& & \multicolumn{4}{c}{MASE/RMSSE} & \multicolumn{4}{c}{MASC/RMSSC} & \multicolumn{4}{c}{MASC\_I/RMSSC\_I} \\
			\cmidrule(lr){3-6} \cmidrule(lr){7-10} \cmidrule(lr){11-14}
			& & M4 & M3 & Favorita  & M5 & M4 & M3 & Favorita  & M5 & M4 & M3 & Favorita  & M5  \\
      \hline
			 \addlinespace

N-BEATS & Base & \textbf{0.726} & \textbf{0.675} & - & - & 0.104 & 0.089 & - & - & 0.313 & 0.235 & - & - \\ 
\cmidrule(lr){2-14}
   & PI\_0.2 & \textbf{\textit{0.727}} & \textbf{\textit{0.676}} & - & - & 0.091 & 0.076 & - & - & 0.297 & 0.221 & - & - \\ 
   & PI\_0.4 & 0.729 & \textbf{\textit{0.676}} & - & - & 0.084 & 0.069 & - & - & 0.280 & 0.207 & - & - \\ 
   & PI\_0.5 & 0.731 & \textbf{\textit{0.676}} & - & - & \textbf{\textit{0.082}} & \textbf{\textit{0.067}} & - & - & 0.273 & 0.201 & - & - \\ 
   & PI\_0.6 & 0.732 & \textbf{\textit{0.676}} & - & - & \textbf{\textit{0.082}} & \textbf{\textit{0.067}} & - & - & 0.265 & 0.196 & - & - \\ 
   & PI\_0.8 & 0.737 & 0.677 & - & - & 0.084 & 0.069 & - & - & 0.250 & 0.186 & - & - \\ 
   & PI\_1 & 0.741 & 0.678 & - & - & 0.090 & 0.075 & - & - & \textbf{\textit{0.235}} & \textbf{\textit{0.179}} & - & - \\ 
   \cmidrule(lr){2-14}
   & FI\_0.2 & \textbf{\textit{0.727}} & \textbf{\textit{0.676}} & - & - & 0.090 & 0.075 & - & - & 0.293 & 0.218 & - & - \\ 
   & FI\_0.4 & 0.729 & \textbf{\textit{0.676}} & - & - & 0.077 & 0.062 & - & - & 0.261 & 0.192 & - & - \\ 
   & FI\_0.5 & 0.731 & \textbf{\textit{0.676}} & - & - & 0.070 & 0.055 & - & - & 0.238 & 0.175 & - & - \\ 
   & FI\_0.6 & 0.733 & \textbf{\textit{0.676}} & - & - & 0.062 & 0.047 & - & - & 0.210 & 0.153 & - & - \\ 
  & FI\_0.8 & 0.740 & 0.677 & - & - & 0.038 & 0.023 & - & - & 0.126 & 0.089 & - & - \\ 
   & FI\_1 & 0.754 & 0.678 & - & - & \textbf{0.000} & \textbf{0.000} & - & - & \textbf{0.000} & \textbf{0.000} & - & - \\ 
            \Xhline{5\arrayrulewidth}
			 \addlinespace

PR & Base & 0.803 & 0.719 & 0.724 & \textbf{1.252} & 0.114 & 0.125 & 0.194 & 0.096 & 0.323 & 0.250 & 0.256 & 0.237 \\ 
\cmidrule(lr){2-14}
   & PI\_0.2 & 0.802 & 0.718 & $\textbf{0.723}^*$ & \textbf{\textit{1.253}} & 0.101 & 0.102 & $0.152^*$ & 0.080 & 0.307 & 0.236 & $0.237^*$ & 0.231 \\ 
   & PI\_0.4 & 0.801 & $\textbf{0.717}^*$ & 0.724 & \textbf{\textit{1.253}} & 0.094 & $0.085^*$ & 0.122 & 0.070 & 0.290 & $0.222^*$ & 0.225 & 0.228 \\ 
   & PI\_0.5 & 0.801 & $\textbf{0.717}^*$ & 0.725 & \textbf{\textit{1.253}} & \textbf{\textit{0.092}} & $\textbf{\textit{0.082}}^*$ & \textbf{\textit{0.115}} & \textbf{\textit{0.069}} & 0.283 & $0.216^*$ & 0.221 & \textbf{\textit{0.227}} \\ 
   & PI\_0.6 & 0.801 & $\textbf{0.717}^*$ & 0.727 & 1.254 & \textbf{\textit{0.092}} & $\textbf{\textit{0.082}}^*$ & 0.116 & 0.070 & 0.275 & $0.211^*$ & \textbf{\textit{0.220}} & \textbf{\textit{0.227}} \\ 
   & PI\_0.8 & 0.801 & 0.718 & 0.731 & 1.255 & 0.094 & 0.090 & 0.136 & 0.080 & 0.260 & 0.201 & 0.224 & 0.228 \\ 
   & PI\_1 & $\textbf{0.800}^*$ & 0.719 & 0.737 & 1.256 & $0.100^*$ & 0.105 & 0.173 & 0.095 & $\textbf{\textit{0.245}}^*$ & \textbf{\textit{0.194}} & 0.236 & 0.231 \\ 
   \cmidrule(lr){2-14}
   & FI\_0.2 & 0.802 & \textbf{\textit{0.718}} & $\textbf{0.723}^*$ & \textbf{\textit{1.253}} & 0.100 & 0.102 & $0.151^*$ & 0.078 & 0.303 & 0.233 & $0.234^*$ & 0.230 \\ 
   & FI\_0.4 & 0.801 & \textbf{\textit{0.718}} & 0.725 & \textbf{\textit{1.253}} & 0.087 & 0.081 & 0.113 & 0.061 & 0.271 & 0.207 & 0.209 & 0.222 \\ 
   & FI\_0.5 & $\textbf{0.800}^*$ & 0.719 & 0.728 & 1.254 & $0.080^*$ & 0.072 & 0.095 & 0.053 & $0.248^*$ & 0.190 & 0.192 & 0.216 \\ 
   & FI\_0.6 & $\textbf{0.800}^*$ & 0.720 & 0.731 & 1.255 & $0.072^*$ & 0.062 & 0.077 & 0.044 & $0.220^*$ & 0.168 & 0.172 & 0.208 \\ 
   & FI\_0.8 & $\textbf{0.800}^*$ & 0.723 & 0.743 & 1.262 & $0.048^*$ & 0.038 & 0.042 & 0.026 & $0.136^*$ & 0.104 & 0.110 & 0.176 \\ 
   & FI\_1 & 0.802 & 0.731 & 0.771 & 1.311 & \textbf{0.000} & \textbf{0.000} & \textbf{0.000} & \textbf{0.000} & \textbf{0.000} & \textbf{0.000} & \textbf{0.000} & \textbf{0.000} \\
			\Xhline{5\arrayrulewidth}
			 \addlinespace

LightGBM & Base & 0.790 & 0.709 & 0.720 & 1.252 & 0.119 & 0.126 & 0.078 & 0.077 & 0.346 & 0.234 & 0.115 & 0.144 \\ 
\cmidrule(lr){2-14}
   & PI\_0.2 & 0.789 & 0.708 & $\textbf{0.719}^*$ & \textbf{\textit{1.251}} & 0.105 & 0.101 & $0.062^*$ & 0.067 & 0.328 & 0.219 & $0.108^*$ & 0.140 \\ 
   & PI\_0.4 & 0.787 & \textbf{\textit{0.707}} & $\textbf{0.719}^*$ & \textbf{\textit{1.251}} & 0.097 & 0.085 & $0.052^*$ & 0.061 & 0.311 & 0.207 & $0.103^*$ & 0.137 \\ 
   & PI\_0.5 & 0.787 & \textbf{\textit{0.707}} & 0.720 & \textbf{\textit{1.251}} & \textbf{\textit{0.095}} & \textbf{\textit{0.082}} & \textbf{\textit{0.050}} & \textbf{\textit{0.060}} & 0.303 & 0.201 & 0.102 & \textbf{\textit{0.136}} \\ 
   & PI\_0.6 & 0.786 & \textbf{\textit{0.707}} & 0.720 & \textbf{\textit{1.251}} & 0.096 & 0.084 & \textbf{\textit{0.050}} & \textbf{\textit{0.060}} & 0.295 & 0.196 & \textbf{\textit{0.101}} & \textbf{\textit{0.136}} \\ 
   & PI\_0.8 & 0.786 & 0.708 & 0.721 & 1.252 & 0.100 & 0.094 & 0.059 & 0.066 & 0.279 & 0.188 & 0.102 & 0.138 \\ 
   & PI\_1 & \textbf{\textit{0.785}} & 0.708 & 0.721 & 1.252 & 0.107 & 0.110 & 0.073 & 0.076 & \textbf{\textit{0.264}} & \textbf{\textit{0.182}} & 0.106 & 0.141 \\ 
   \cmidrule(lr){2-14}
   & FI\_0.2 & 0.788 & 0.708 & \textbf{\textit{0.720}} & 1.251 & 0.104 & 0.100 & 0.062 & 0.065 & 0.324 & 0.216 & 0.106 & 0.138 \\ 
   & FI\_0.4 & 0.786 & 0.707 & \textbf{\textit{0.720}} & 1.251 & 0.090 & 0.079 & 0.047 & 0.051 & 0.291 & 0.192 & 0.096 & 0.130 \\ 
   & FI\_0.5 & 0.785 & 0.707 & 0.721 & $\textbf{1.250}^*$ & 0.084 & 0.069 & 0.040 & $0.044^*$ & 0.267 & 0.176 & 0.088 & $0.125^*$ \\ 
   & FI\_0.6 & 0.783 & $\textbf{0.706}^*$ & 0.722 & $\textbf{1.250}^*$ & 0.076 & $0.059^*$ & 0.033 & $0.036^*$ & 0.236 & $0.155^*$ & 0.079 & $0.119^*$ \\ 
   & FI\_0.8 & 0.780 & 0.708 & 0.726 & 1.251 & 0.050 & 0.036 & 0.019 & 0.019 & 0.146 & 0.095 & 0.051 & 0.098 \\ 
   & FI\_1 & $\textbf{0.778}^*$ & 0.712 & 0.734 & 1.261 & $\textbf{0.000}^*$ & \textbf{0.000} & \textbf{0.000} & \textbf{0.000} & $\textbf{0.000}^*$ & \textbf{0.000} & \textbf{0.000} & \textbf{0.000} \\
            \Xhline{5\arrayrulewidth}
			 \addlinespace

ETS & Base & \textbf{0.749} & \textbf{0.698} & \textbf{0.700} & 1.249 & 0.126 & 0.013 & 0.212 & 0.097 & 0.198 & 0.018 & 0.250 & 0.107 \\ 
\cmidrule(lr){2-14}
   & PI\_0.2 & \textbf{\textit{0.750}} & \textbf{\textit{0.699}} & \textbf{\textit{0.701}} & \textbf{\textit{1.249}} & 0.104 & 0.010 & 0.166 & 0.078 & 0.184 & 0.016 & 0.227 & 0.098 \\ 
   & PI\_0.4 & 0.752 & \textbf{\textit{0.699}} & 0.705 & \textbf{\textit{1.249}} & 0.089 & 0.008 & 0.132 & 0.065 & 0.172 & 0.016 & 0.211 & 0.093 \\ 
   & PI\_0.5 & 0.754 & \textbf{\textit{0.699}} & 0.708 & \textbf{\textit{1.249}} & \textbf{\textit{0.085}} & \textbf{\textit{0.007}} & 0.124 & \textbf{\textit{0.063}} & 0.167 & 0.015 & 0.206 & \textbf{\textit{0.092}} \\ 
   & PI\_0.6 & 0.755 & \textbf{\textit{0.699}} & 0.711 & \textbf{\textit{1.249}} & \textbf{\textit{0.085}} & 0.008 & \textbf{\textit{0.123}} & 0.064 & 0.163 & 0.015 & \textbf{\textit{0.204}} & 0.093 \\ 
   & PI\_0.8 & 0.760 & 0.700 & 0.720 & 1.250 & 0.090 & 0.009 & 0.145 & 0.075 & 0.156 & \textbf{\textit{0.014}} & 0.208 & 0.096 \\ 
   & PI\_1 & 0.764 & 0.701 & 0.732 & 1.252 & 0.101 & 0.010 & 0.184 & 0.094 & \textbf{\textit{0.152}} & \textbf{\textit{0.014}} & 0.221 & 0.104 \\ 
   \cmidrule(lr){2-14}
   & FI\_0.2 & \textbf{\textit{0.750}} & \textbf{\textit{0.699}} & \textbf{\textit{0.701}} & 1.249 & 0.103 & 0.010 & 0.166 & 0.076 & 0.181 & 0.016 & 0.224 & 0.097 \\ 
   & FI\_0.4 & 0.752 & \textbf{\textit{0.699}} & 0.704 & $\textbf{1.248}^*$ & 0.081 & 0.008 & 0.124 & $0.057^*$ & 0.159 & 0.014 & 0.193 & $0.086^*$ \\ 
   & FI\_0.5 & 0.754 & \textbf{\textit{0.699}} & 0.707 & $\textbf{1.248}^*$ & 0.070 & 0.006 & 0.104 & $0.048^*$ & 0.145 & 0.013 & 0.174 & $0.080^*$ \\ 
   & FI\_0.6 & 0.756 & \textbf{\textit{0.699}} & 0.710 & $\textbf{1.248}^*$ & 0.059 & 0.005 & 0.084 & $0.038^*$ & 0.127 & 0.012 & 0.152 & $0.073^*$ \\ 
   & FI\_0.8 & 0.763 & 0.700 & 0.721 & 1.249 & 0.033 & 0.003 & 0.044 & 0.019 & 0.078 & 0.007 & 0.093 & 0.056 \\ 
   & FI\_1 & 0.777 & 0.701 & 0.743 & 1.250 & \textbf{0.000} & \textbf{0.000} & \textbf{0.000} & \textbf{0.000} & \textbf{0.000} & \textbf{0.000} & \textbf{0.000} & \textbf{0.000} \\
            \Xhline{5\arrayrulewidth}
			 \addlinespace

ARIMA & Base & 0.777 & 0.712 & \textbf{0.734} & \textbf{1.258} & 0.308 & 0.196 & 0.178 & 0.150 & 0.581 & 0.320 & 0.325 & 0.417 \\ 
\cmidrule(lr){2-14}
   & PI\_0.2 & \textbf{\textit{0.775}} & 0.710 & \textbf{\textit{0.738}} & \textbf{\textit{1.259}} & 0.256 & 0.156 & 0.147 & 0.124 & 0.545 & 0.297 & 0.309 & 0.408 \\ 
   & PI\_0.4 & \textbf{\textit{0.775}} & \textbf{\textit{0.709}} & 0.744 & 1.261 & 0.221 & 0.128 & 0.125 & 0.108 & 0.514 & 0.278 & 0.297 & 0.402 \\ 
   & PI\_0.5 & \textbf{\textit{0.775}} & \textbf{\textit{0.709}} & 0.747 & 1.262 & \textbf{\textit{0.213}} & \textbf{\textit{0.122}} & \textbf{\textit{0.120}} & \textbf{\textit{0.106}} & 0.500 & 0.271 & 0.293 & 0.400 \\ 
   & PI\_0.6 & 0.776 & 0.710 & 0.750 & 1.263 & \textbf{\textit{0.213}} & 0.123 & 0.121 & 0.108 & 0.487 & 0.264 & \textbf{\textit{0.291}} & \textbf{\textit{0.399}} \\ 
   & PI\_0.8 & 0.779 & 0.712 & 0.759 & 1.266 & 0.228 & 0.138 & 0.137 & 0.124 & 0.466 & 0.254 & \textbf{\textit{0.291}} & 0.400 \\ 
   & PI\_1 & 0.783 & 0.714 & 0.768 & 1.269 & 0.257 & 0.163 & 0.166 & 0.149 & \textbf{\textit{0.450}} & \textbf{\textit{0.248}} & 0.296 & 0.405 \\ 
   \cmidrule(lr){2-14}
   & FI\_0.2 & 0.774 & 0.710 & \textbf{\textit{0.739}} & \textbf{\textit{1.259}} & 0.253 & 0.156 & 0.145 & 0.123 & 0.538 & 0.293 & 0.305 & 0.406 \\ 
   & FI\_0.4 & 0.772 & 0.708 & 0.748 & 1.262 & 0.203 & 0.120 & 0.114 & 0.097 & 0.478 & 0.260 & 0.277 & 0.392 \\ 
   & FI\_0.5 & $\textbf{0.771}^*$ & 0.707 & 0.756 & 1.265 & $0.178^*$ & 0.103 & 0.099 & 0.084 & $0.438^*$ & 0.238 & 0.258 & 0.382 \\ 
   & FI\_0.6 & $\textbf{0.771}^*$ & $\textbf{0.706}^*$ & 0.765 & 1.270 & $0.152^*$ & $0.087^*$ & 0.084 & 0.071 & $0.387^*$ & $0.210^*$ & 0.232 & 0.369 \\ 
   & FI\_0.8 & $\textbf{0.771}^*$ & $\textbf{0.706}^*$ & 0.796 & 1.295 & $0.091^*$ & $0.050^*$ & 0.049 & 0.043 & $0.240^*$ & $0.131^*$ & 0.150 & 0.311 \\ 
   & FI\_1 & 0.779 & 0.710 & 0.857 & 1.445 & \textbf{0.000} & \textbf{0.000} & \textbf{0.000} & \textbf{0.000} & \textbf{0.000} & \textbf{0.000} & \textbf{0.000} & \textbf{0.000} \\
\Xhline{5\arrayrulewidth}
            
			\end{tabular}
		}	
		\label{tab:hor_results}
\end{center}
\end{table}

\subsection{Results of Horizontal Stability Experiments}

As outlined earlier, for horizontal stability, it is usually desirable to not stabilise towards a flat line but to take into account known, obvious patterns in the series. Following this idea, in our horizontal stability experiments, we perform a decomposition of the series, and then focus for prediction and stabilisation with our methods on the remainders.
Although this renders the stabilisation effectively not model-agnostic anymore, it has the advantage that the stabilisation is straightforward and comparable to the vertical case. In particular, we use MSTL decomposition~\citep{Bandara2022MSTL} and the STL forecast function \verb|stlf| \citep{ref_22} where seasonality is forecast by repeating the last seasonal period into the future, and the trend is forecast using an ETS/ZZN model \citep{ref_4}.

Table \ref{tab:hor_results} shows the results of the horizontal stability experiments for all experimental datasets. 
Accuracy is reported for the full series against actuals, whereas stability is reported for the remainders of the seasonal decomposition. 
If we evaluate stability over the full series, it would be dominated by the volatility of seasonal and trend patterns, and not by the remainder over which we actually stabilise. This makes it difficult to compare across datasets and could lead to misleading interpretations.
For the M4 and M3 datasets, the results are reported for MASE, MASC, and MASC\_I, whereas for the Favorita and M5 datasets, the results are reported for RMSSE, RMSSC, and RMSSC\_I. 
The results in Table \ref{tab:hor_results} are grouped in the same way as in Table \ref{tab:ver_results}, with analogous highlighting.

In terms of accuracy, the base models outperform the proposed partial and full interpolation models on the M3 and M4 datasets for N-BEATS, 
on the M5 dataset for PR, on the M3, M4, and Favorita datasets for ETS, and on the Favorita and M5 datasets for ARIMA. For the other cases, especially for the M4 dataset, we can see that there are cases where even the most stable methods are also the most accurate. We hypothesise that this may be due to the following reasons. The M3 and M4 datasets have relatively strong trend and seasonality \citep{godahewa_2021_monash} and are compilations of very different series, such that, for example, the M4 winning method modelled the seasonality locally for each series separately \citep{ref_1}.
Depending on the series, if the signal-to-noise ratio is low in the remainder after modelling the seasonality and trend with a seasonal decomposition approach, the remainder may contain very little useful information. Therefore, in these cases, mostly ignoring the remainder by making it maximally stable into a flat line may yield the best results.

In terms of stability, further to the extreme cases where the maximally stable models are also the most accurate, the proposed models are always better than the base models for all measures used. 
The comparison between the partial and full interpolation models for horizontal stability is similar to the corresponding observations from the vertical stability experiments. Overall, the full interpolation models consistently outperform the partial interpolation models in terms of stability. The full interpolation models use a smoothed previous forecast to obtain the next forecast and thus, provide more stable forecasts compared to the partial interpolation models. In terms of accuracy, the partial and full interpolation models show mixed performances. They are usually very close, with only the most stable FI models occasionally degrading in accuracy. Otherwise, there is no clear winner between the two accuracy-wise.

\subsection{Visualisation of Increased Stability}

\begin{figure}
\centering
  \includegraphics[scale=0.5]{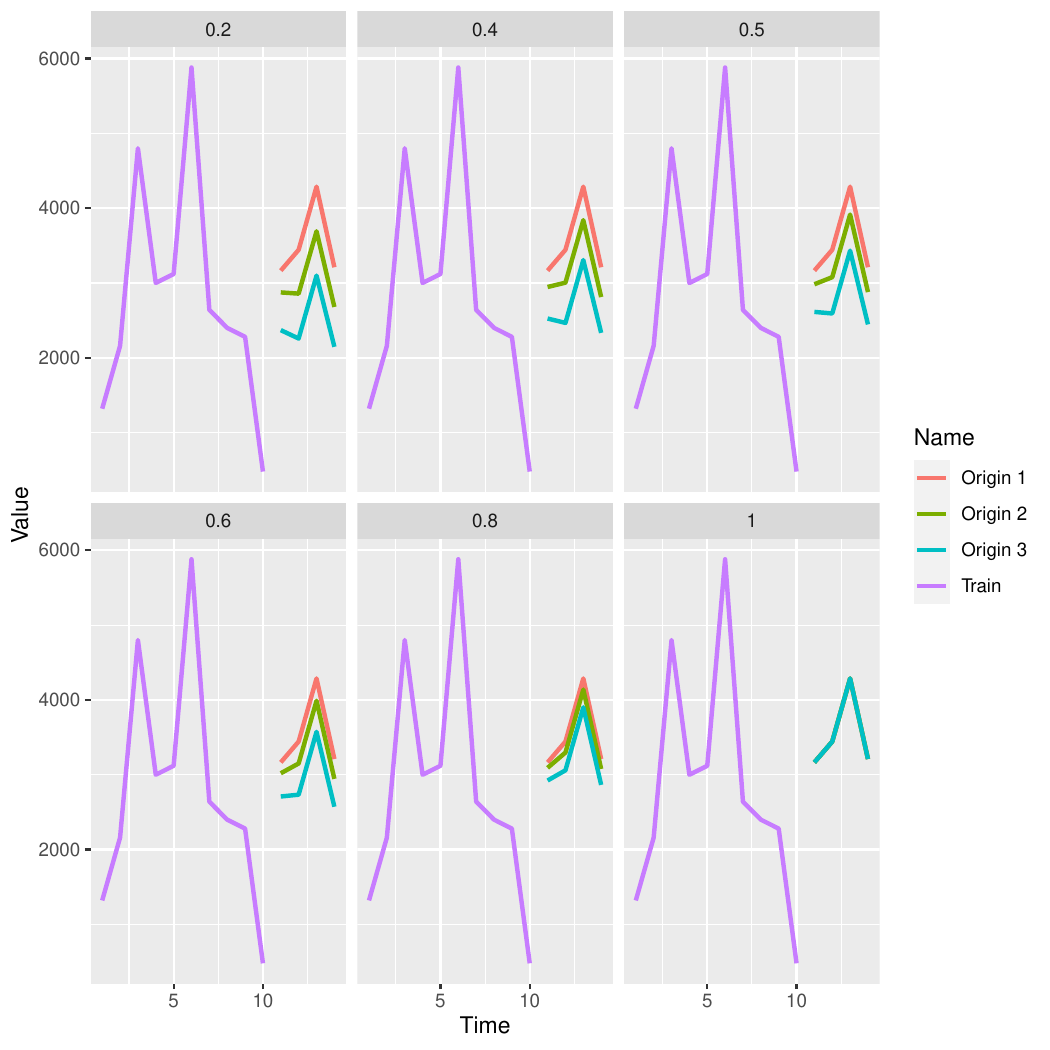}
  \caption{A visualisation of the increase of vertical stability across series N1402 in the M3 monthly dataset for N-BEATS. Each subplot is corresponding with a different w\_s value.}
  \label{fig: m3_nbeats_stability}
\end{figure}

\begin{figure}
\centering
  \includegraphics[scale=0.5]{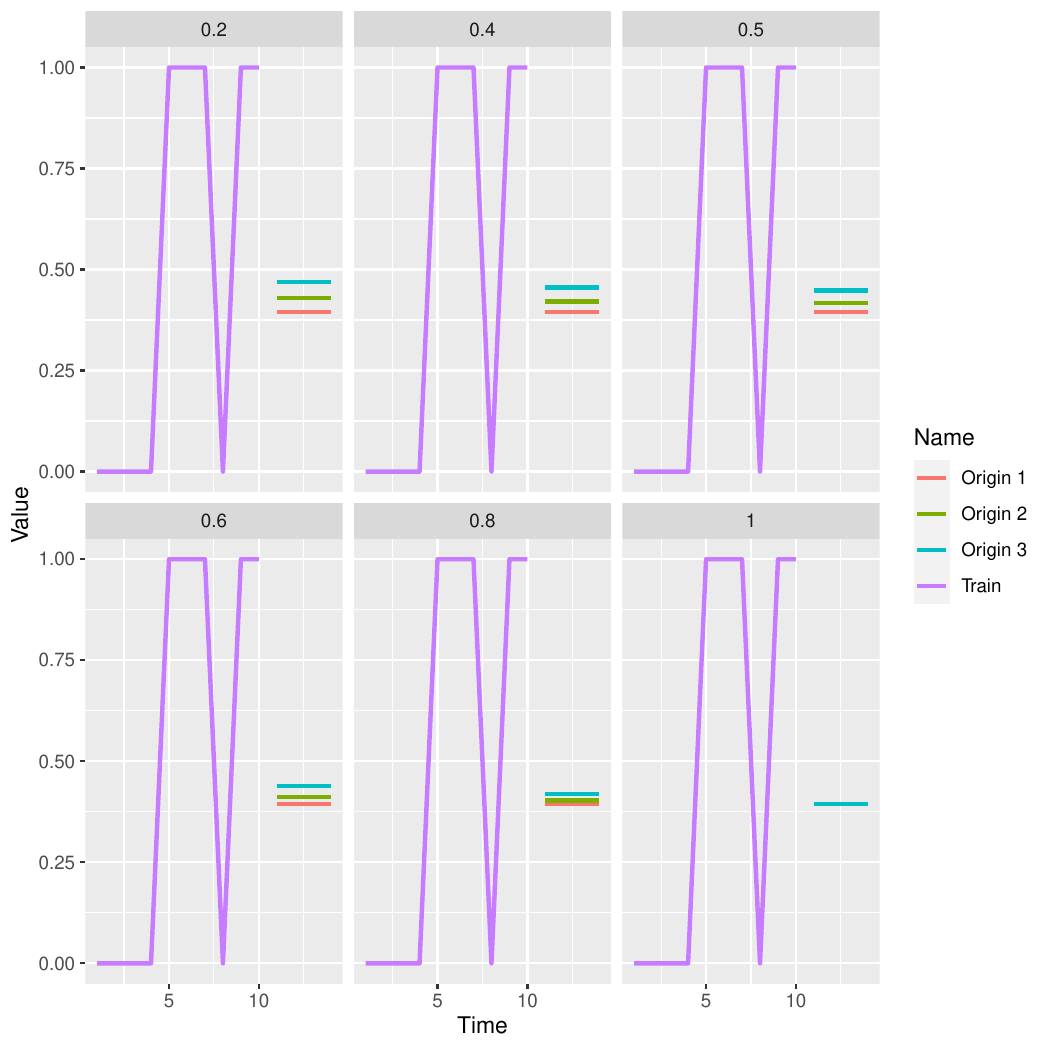}
  \caption{A visualisation of the increase of vertical stability across the $5^{th}$ series in the Favorita dataset for ETS. Each subplot is corresponding with a different w\_s value.}
  \label{fig: favorita_ets_stability}
\end{figure}

\begin{figure}
\centering
  \includegraphics[scale=0.5]{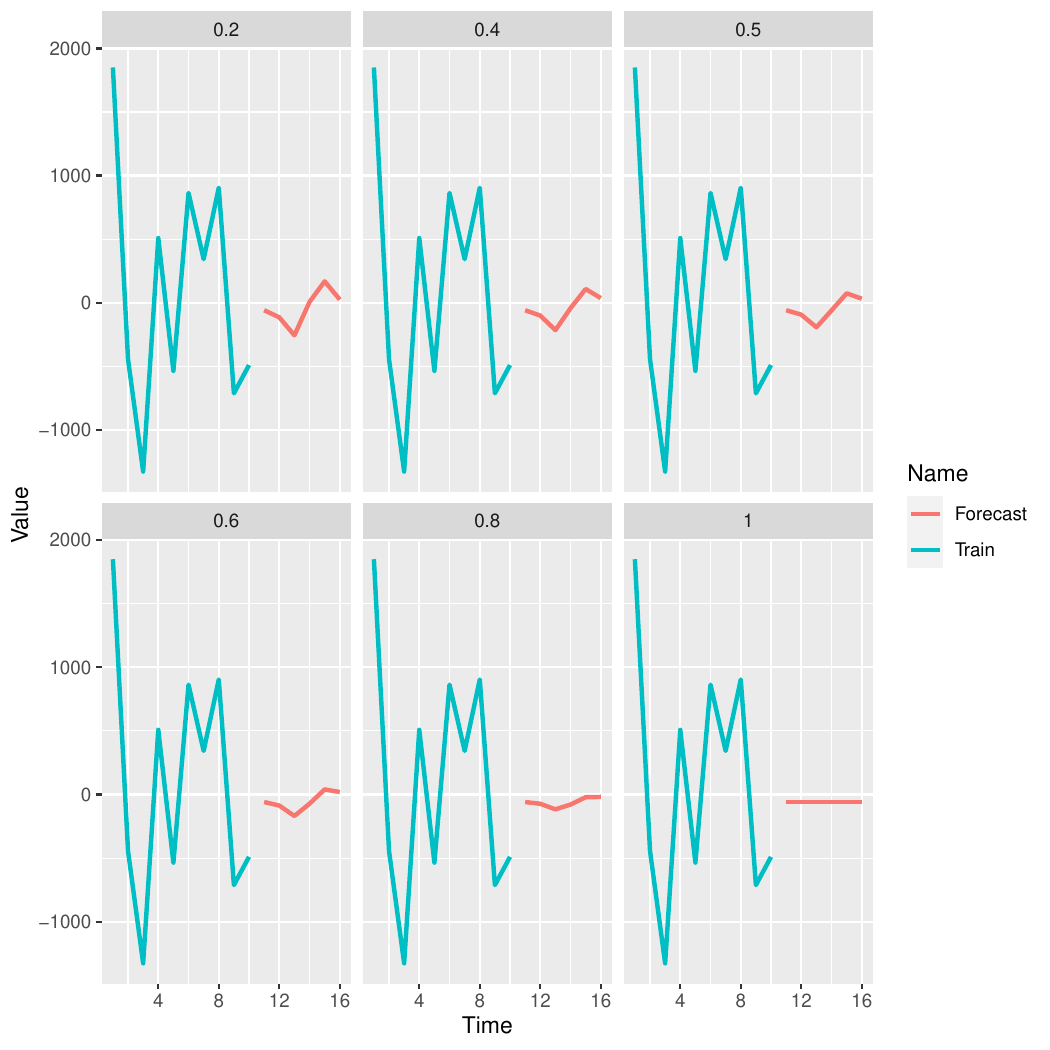}
  \caption{A visualisation of the increase of horizontal stability across the remainder of series N1403 (origin 1) in the M3 monthly dataset for LightGBM. Each subplot is corresponding with a different w\_s value.}
  \label{fig: m3_lightgbm_stability}
\end{figure}

\begin{figure}
\centering
  \includegraphics[scale=0.5]{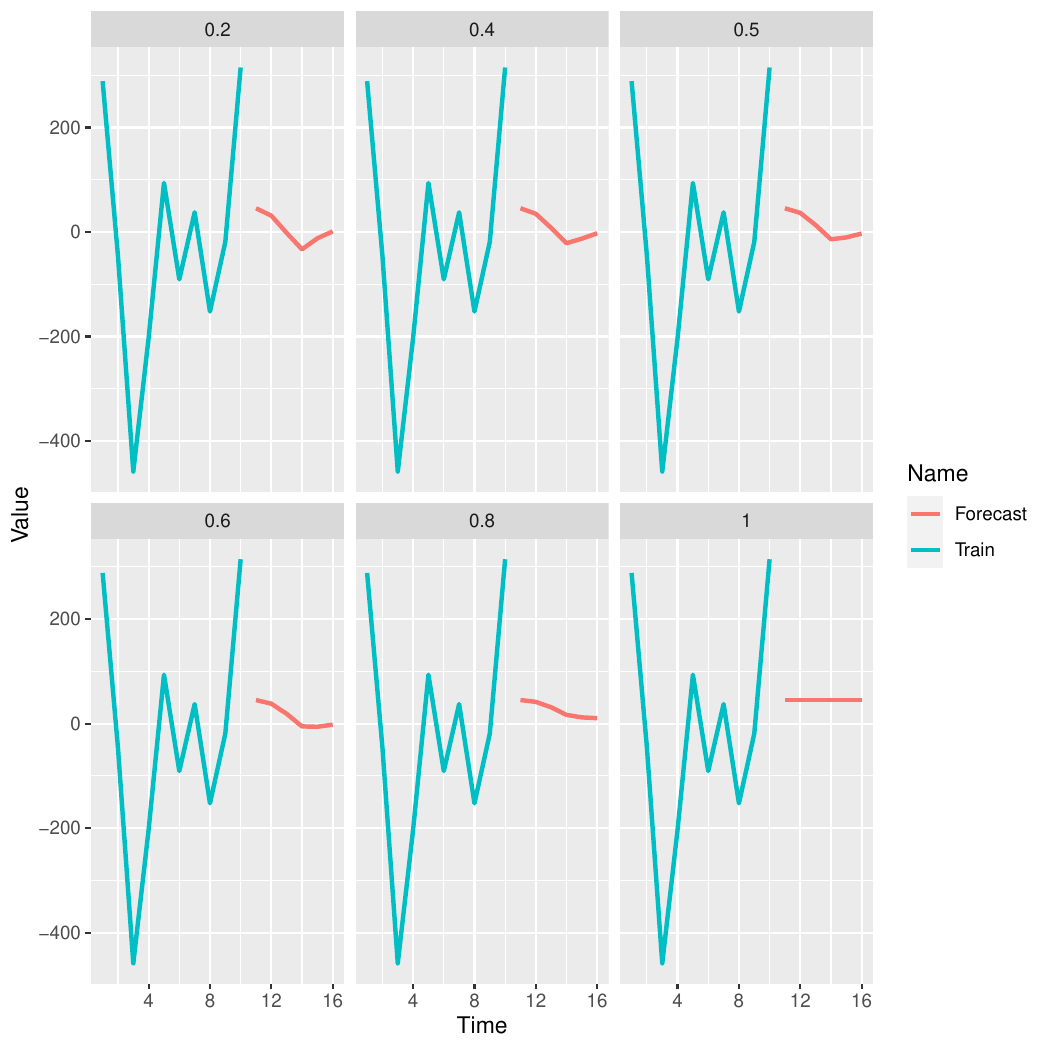}
  \caption{A visualisation of the increase of horizontal stability across the remainder of the series named M2 (origin 1) in the M4 monthly dataset for PR. Each subplot is corresponding with a different w\_s value.}
  \label{fig: m4_pr_stability}
\end{figure}

We illustrate the behaviour of the proposed methodology using examples. Figure~\ref{fig: m3_nbeats_stability} shows an example of N-BEATS for the M3 dataset, and Figure~\ref{fig: favorita_ets_stability} shows an example of ETS for the Favorita dataset. We see that the method works as intended, by bringing the predictions from subsequent origins the closer to the predictions of the first origin the higher $w\_s$ is chosen.
For horizontal stability, we stabilise the remainders after MSTL decomposition. Figures~\ref{fig: m3_lightgbm_stability} and \ref{fig: m4_pr_stability} show examples of a LightGBM prediction for the remainder of a series from the M3 dataset and a prediction from PR for the remainder of a series from the M4 dataset, respectively.

\subsection{Statistical Testing Results}
\label{sec:stat_results}

For statistical testing, we focus on the cases where some of our proposed model variants provide both more accurate and stable forecasts compared with the base models. 
For this, the most accurate interpolation variants are compared separately with the corresponding base models for every series, averaging their corresponding RMSSE, RMSSC and RMSSC\_I errors over the different forecast origins, as we assume that forecasts for the same series from different origins are not independent of each other.

Statistical testing is conducted separately for vertical and horizontal stability models. For the statistical testing of vertical stability models, the models that perform best in terms of accuracy, if different from the base model, are considered, e.g., FI\_0.4 and FI\_0.5 against the PR base model on Favorita, FI\_0.6 and FI\_0.8 against PR base model on M5, and others. 
In the same way, for statistical testing of horizontal stability models, e.g., PI\_0.2 versus PR base model on Favorita, PI\_0.2 and PI\_0.4 versus LightGBM base model on Favorita, and others, are considered.

As explained in Section \ref{sec:statistical_testing}, a Bonferroni correction is applied for the significance level, $\alpha$ by dividing its initial value (0.05) by the total number of comparisons made (102). Thus, the final value of $\alpha$ considered for statistical testing is 0.000490196.

In Tables \ref{tab:ver_results} and \ref{tab:hor_results}, asterisks represent the results that are significantly more accurate and stable than the corresponding base model results. For all pairwise statistical comparisons, the results are highly significant ($p$-value$\ < 10^{-6}$). 

\subsection{Trade-off Analysis with Pareto Fronts}

We effectively perform a multi-objective optimisation in which both stability and accuracy are optimised. For multi-objective optimisation problems, there are different solutions based on the requirements of the users, e.g., solutions with higher accuracy and less stability, solutions with higher stability and less accuracy, and solutions with the same accuracy but more stability than others. A Pareto front is a plot that visualises all the solutions that do not dominate each other.

\begin{figure}
  \centering
  \subfloat[N-BEATS]{\includegraphics[height=0.4\textheight]{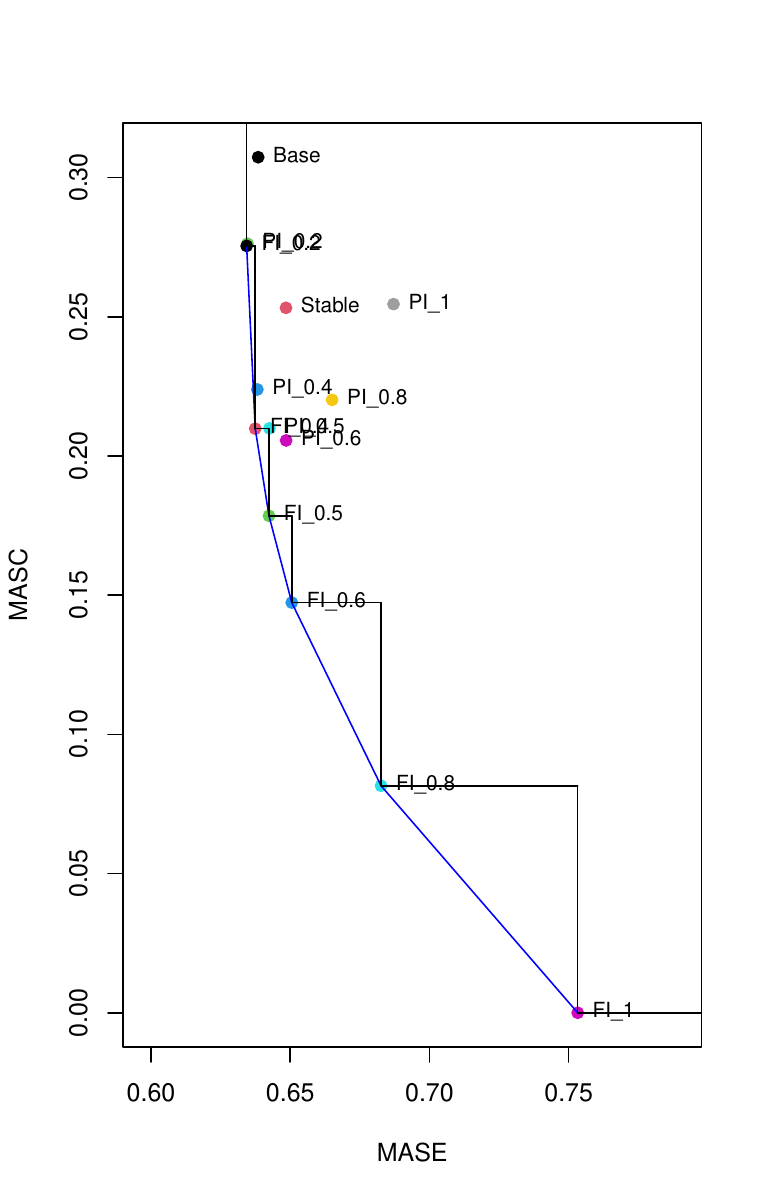}}
  \subfloat[LightGBM]{\includegraphics[height=0.4\textheight]{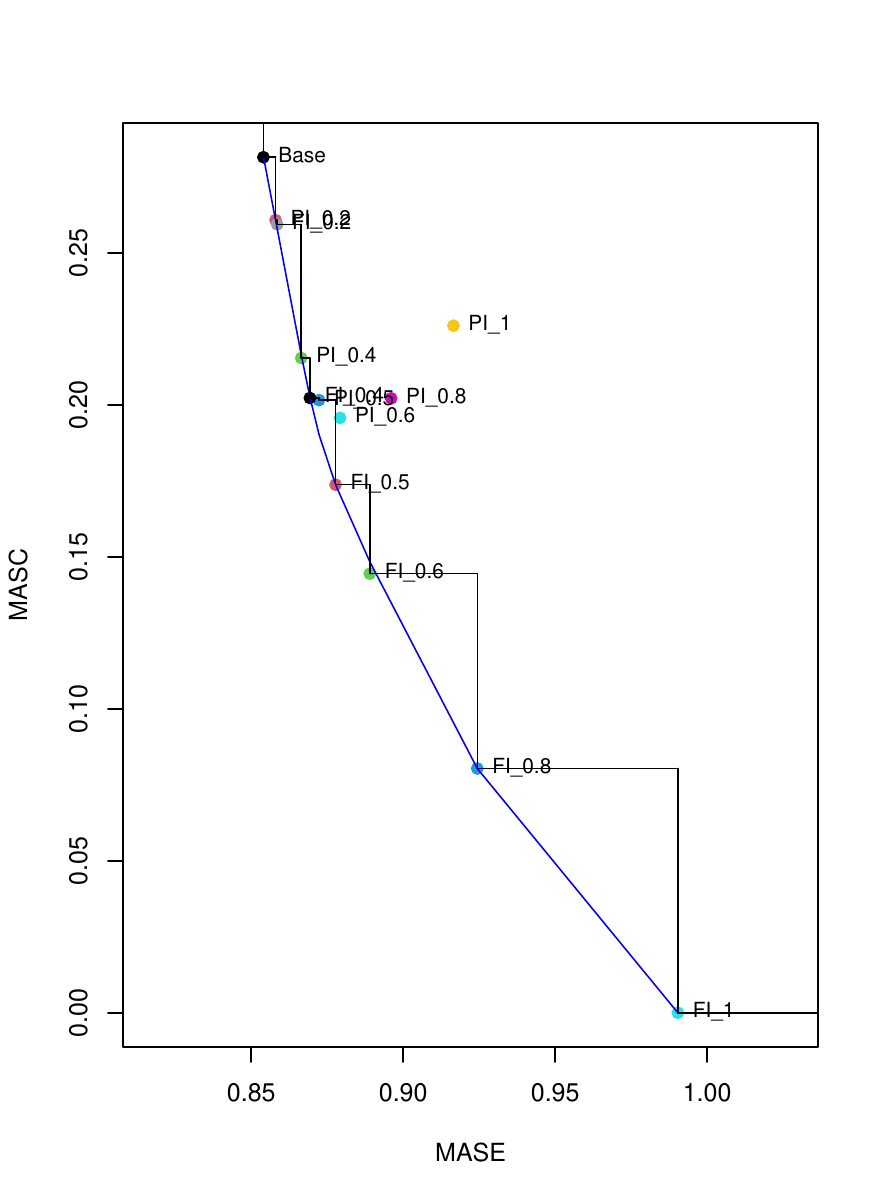}}
    \subfloat[Pooled Regression]{\includegraphics[height=0.4\textheight]{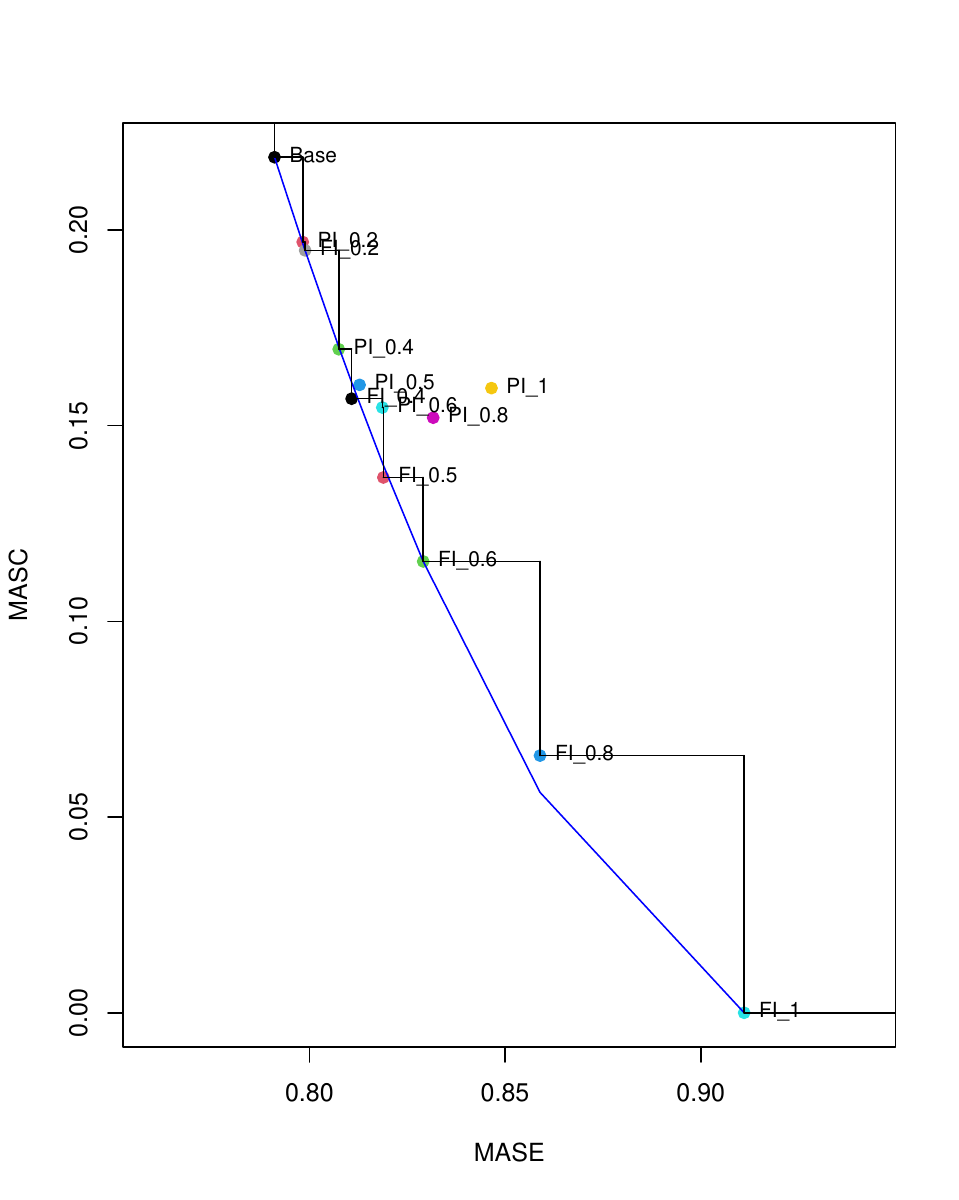}}
      \hfill
  \subfloat[ETS]{\includegraphics[height=0.4\textheight]{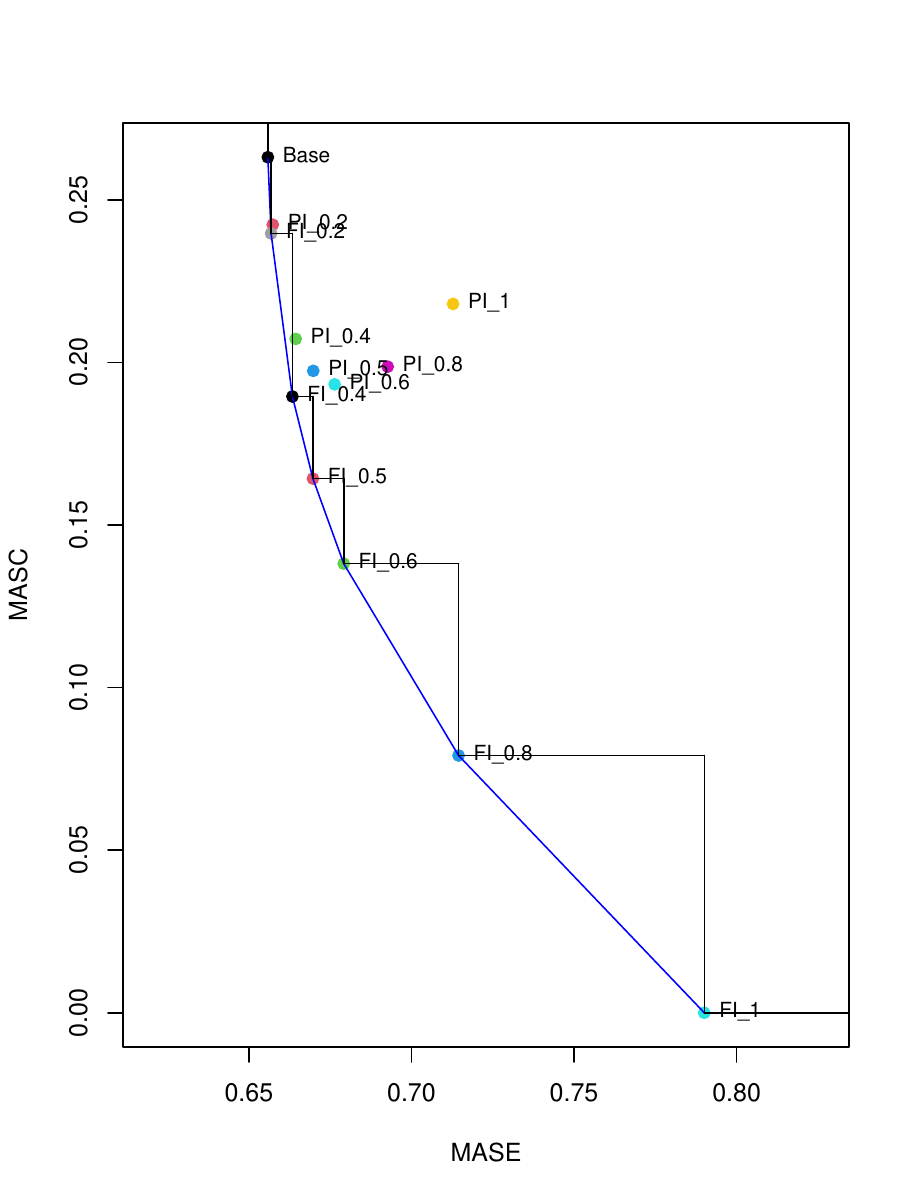}}
  \subfloat[ARIMA]{\includegraphics[height=0.4\textheight]{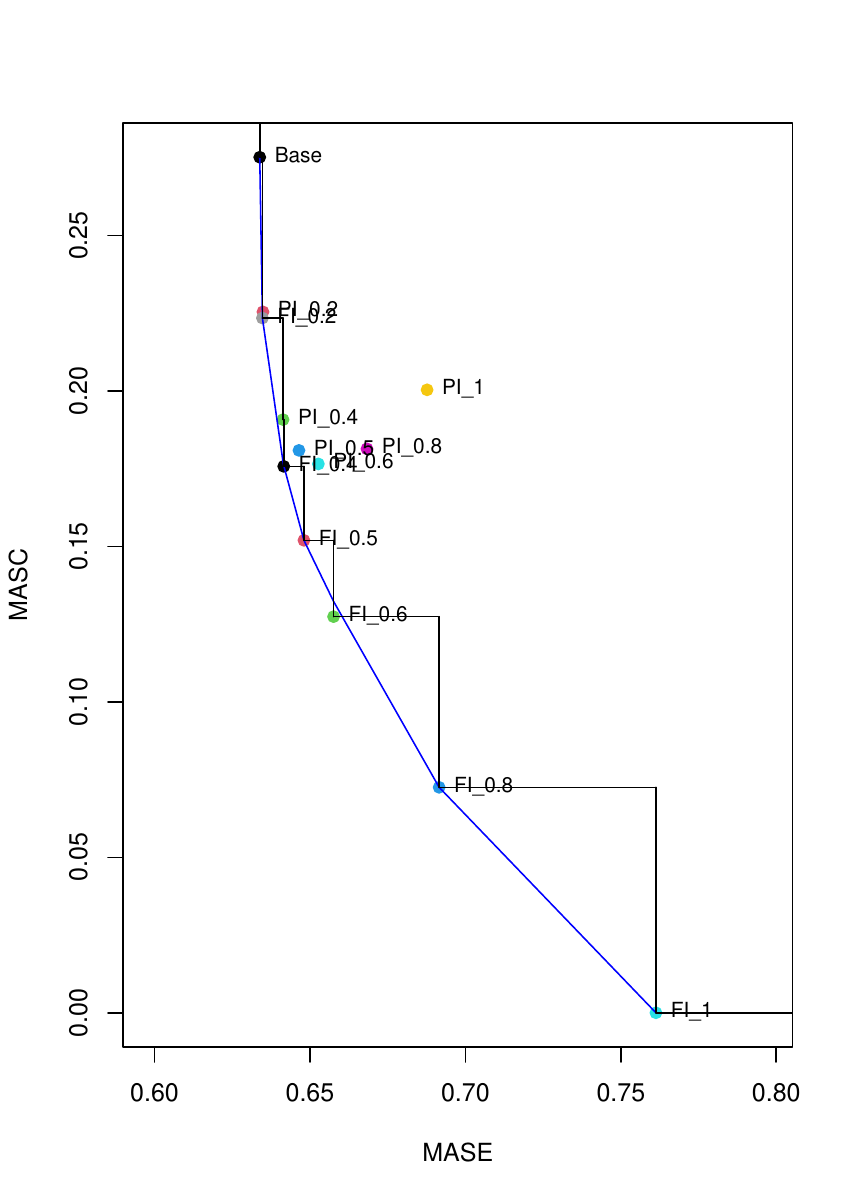}}
  \caption{Pareto fronts showing the accuracy (MASE) and stability (MASC) of all vertical stability experimental models across the M4 dataset. The blue lines show smoothed versions of the Pareto front with COBS splines, for cases where more than 3 solutions are on the Pareto front. The $x$ and $y$ axes use the same scale, so that a steep declining slope indicates a higher gain of stability (decrease of MASC) than what is lost in accuracy (increase of MASE).}
  \label{fig:v-M4}
\end{figure}

\begin{figure}
  \centering
  \subfloat[LightGBM]{\includegraphics[height=0.4\textheight]{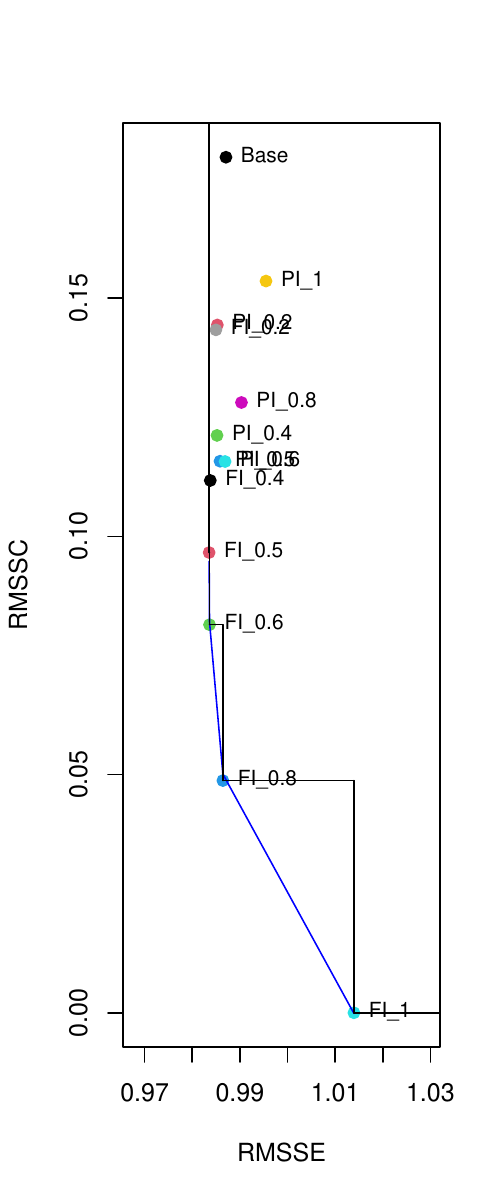}}
  \subfloat[Pooled Regression]{\includegraphics[height=0.4\textheight]{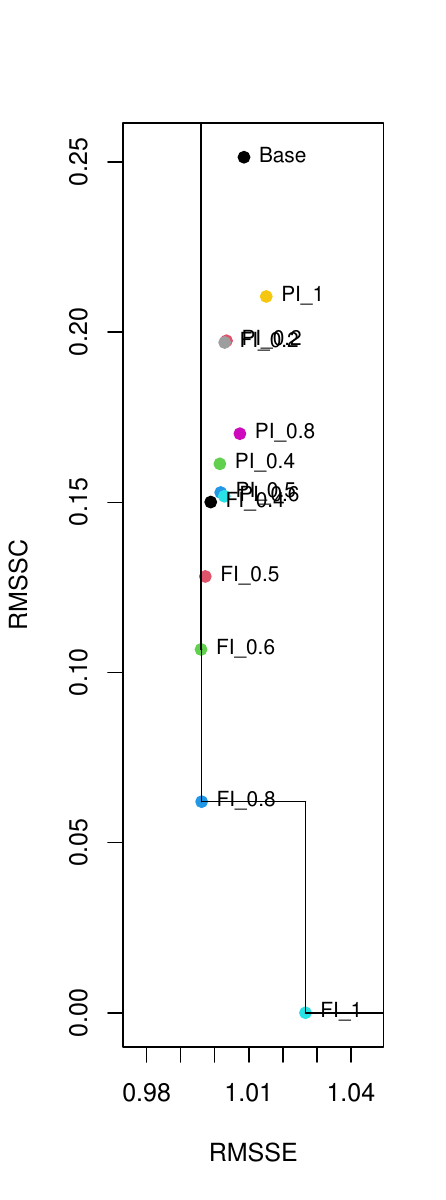}}
  \subfloat[ETS]{\includegraphics[height=0.4\textheight]{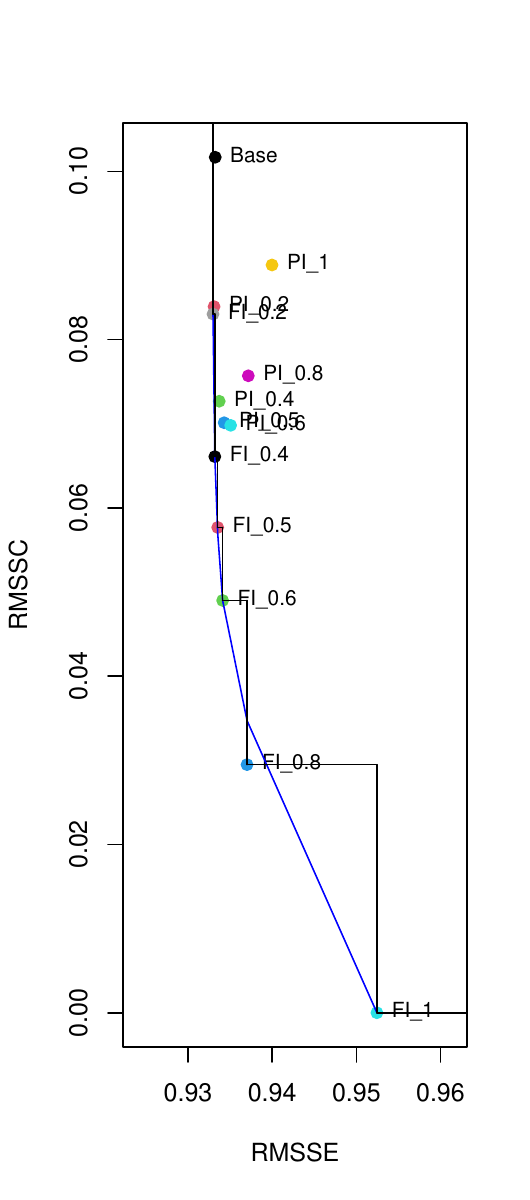}}
      
  \subfloat[ARIMA]{\includegraphics[height=0.4\textheight]{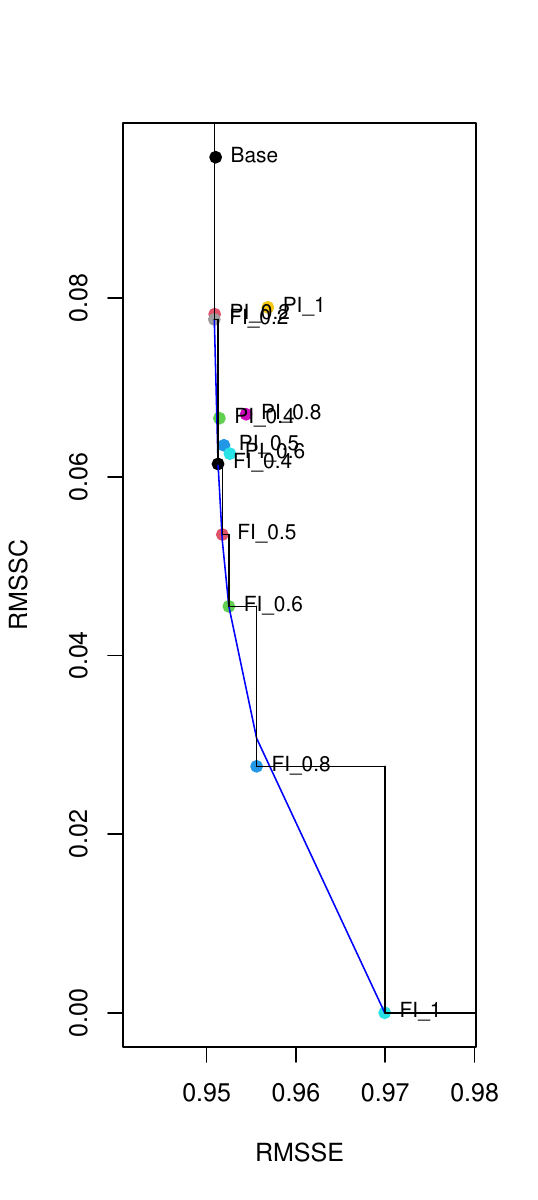}}
  \caption{Pareto fronts showing the accuracy (RMSSE) and stability (RMSSC) of all vertical stability
experimental models across the M5 dataset. The blue lines show smoothed versions of the Pareto front with COBS splines, for cases where more than 3 solutions are on the Pareto front. The $x$ and $y$ axes use the same scale, so that a steep declining slope indicates a higher gain of stability (decrease of RMSSC) than what is lost in accuracy (increase of RMSSE).}
  \label{fig:v-M5}
\end{figure}

\begin{figure}
  \centering
  \subfloat[N-BEATS]{\includegraphics[height=0.4\textheight]{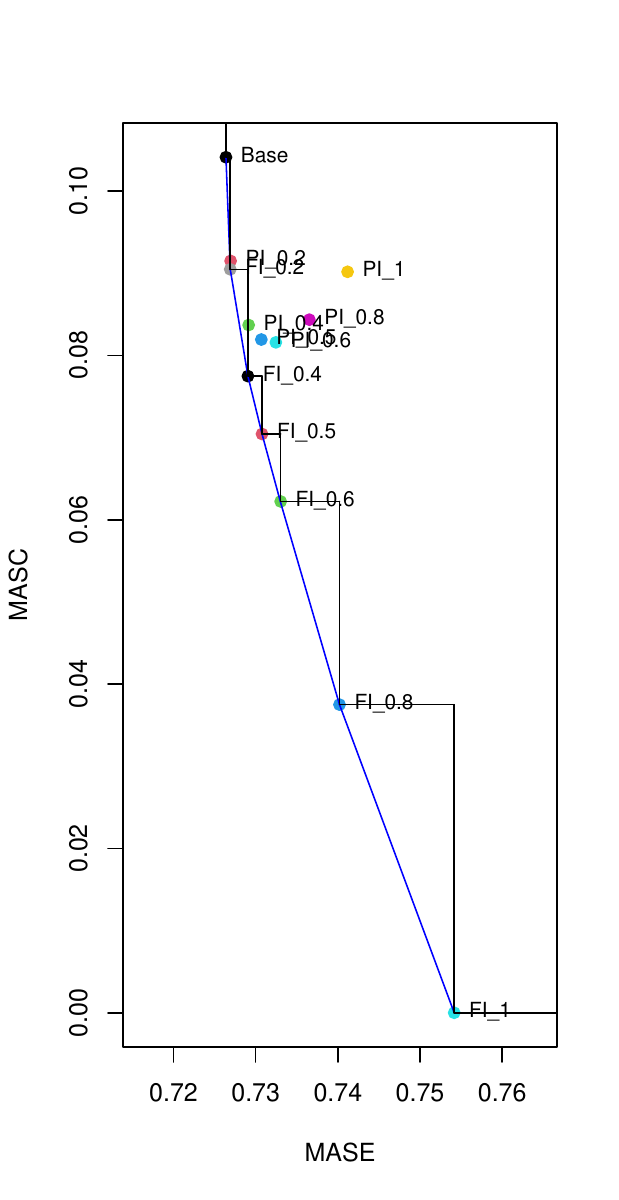}}
  \subfloat[LightGBM]{\includegraphics[height=0.4\textheight]{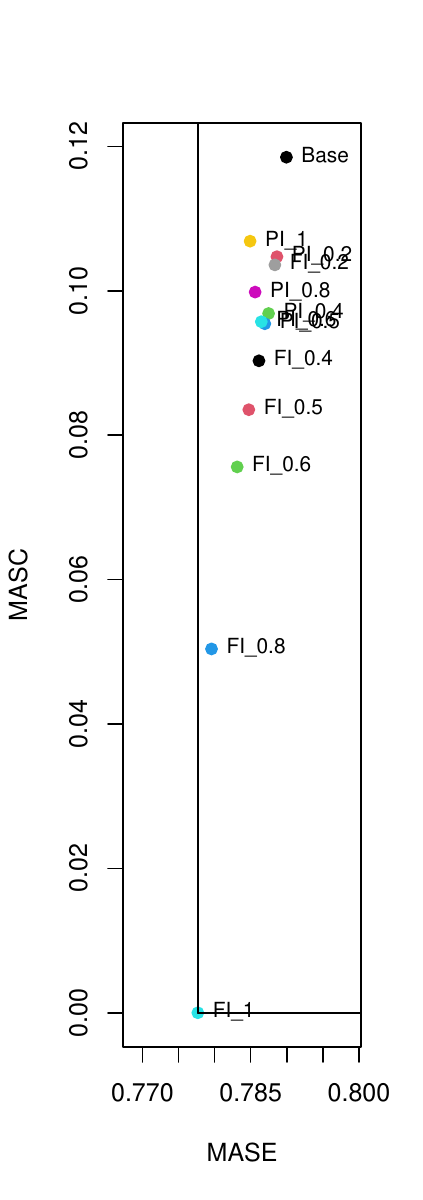}}
    \subfloat[Pooled Regression]{\includegraphics[height=0.4\textheight]{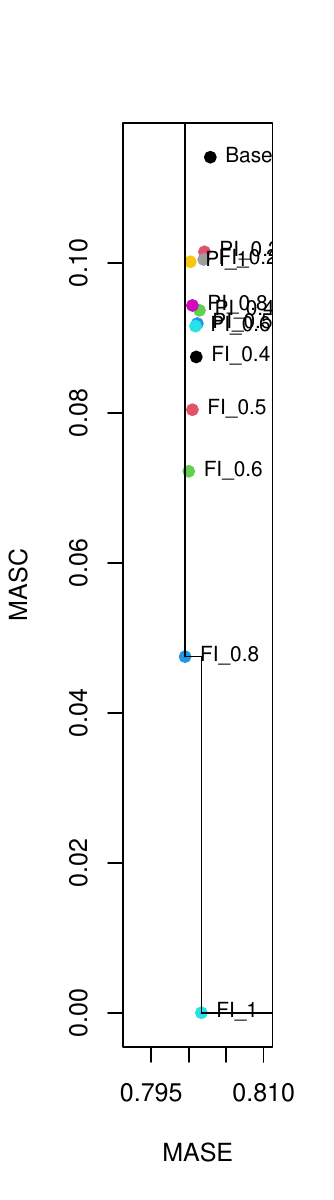}}
      \hfill
  \subfloat[ETS]{\includegraphics[height=0.4\textheight]{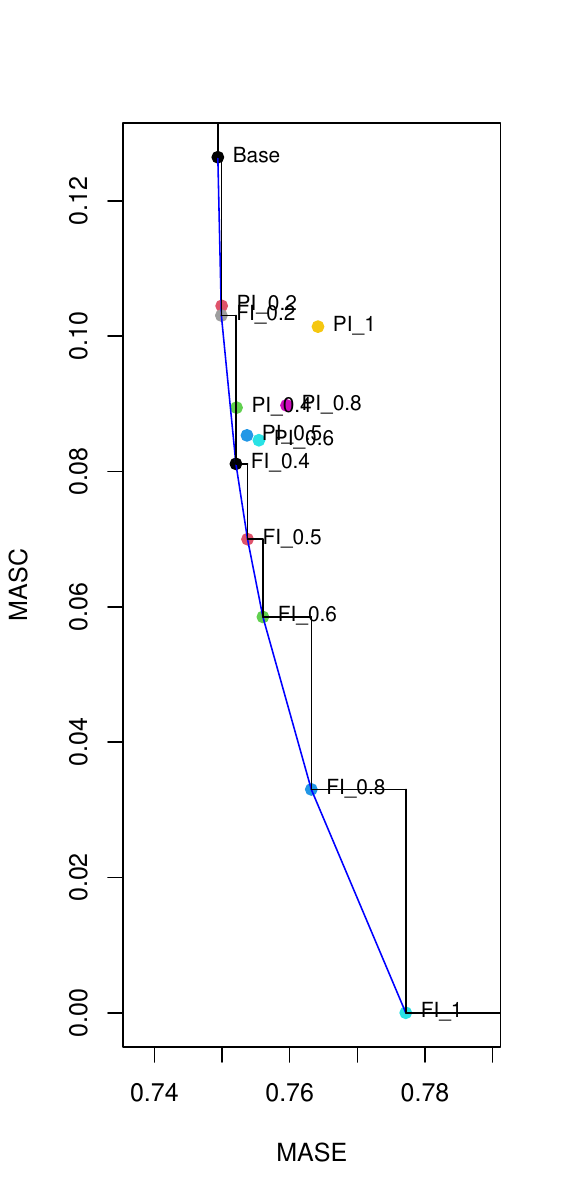}}
  \subfloat[ARIMA]{\includegraphics[height=0.4\textheight]{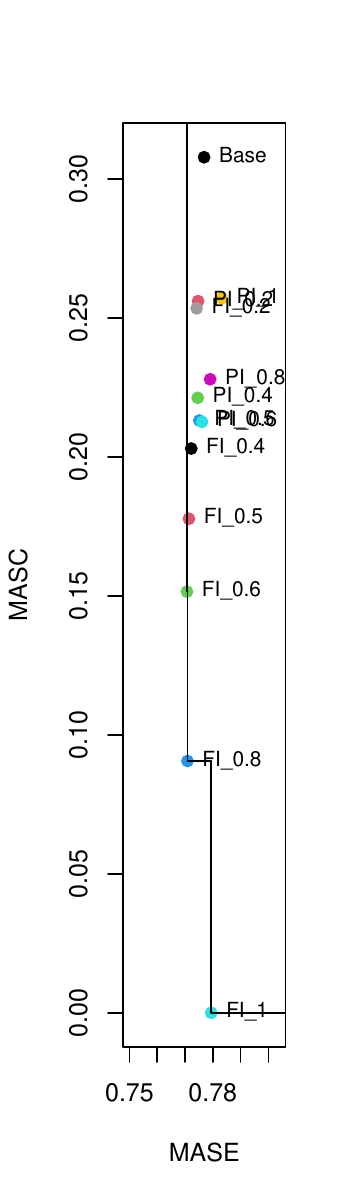}}
  \caption{Pareto fronts showing the accuracy (MASE) and stability (MASC) of all horizontal stability
experimental models across the M4 dataset. The blue lines show smoothed versions of the Pareto front with COBS splines, for cases where more than 3 solutions are on the Pareto front. The $x$ and $y$ axes use the same scale, so that a steep declining slope indicates a higher gain of stability (decrease of MASC) than what is lost in accuracy (increase of MASE).}
  \label{fig:h-M4}
\end{figure}

\begin{figure}
  \centering
  \subfloat[LightGBM]{\includegraphics[height=0.4\textheight]{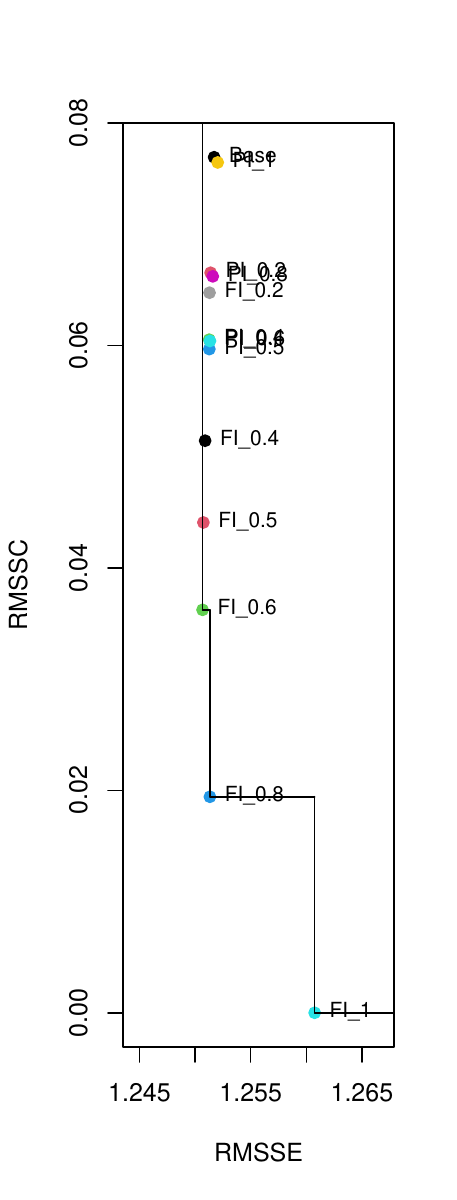}}
    \subfloat[Pooled Regression]{\includegraphics[height=0.4\textheight]{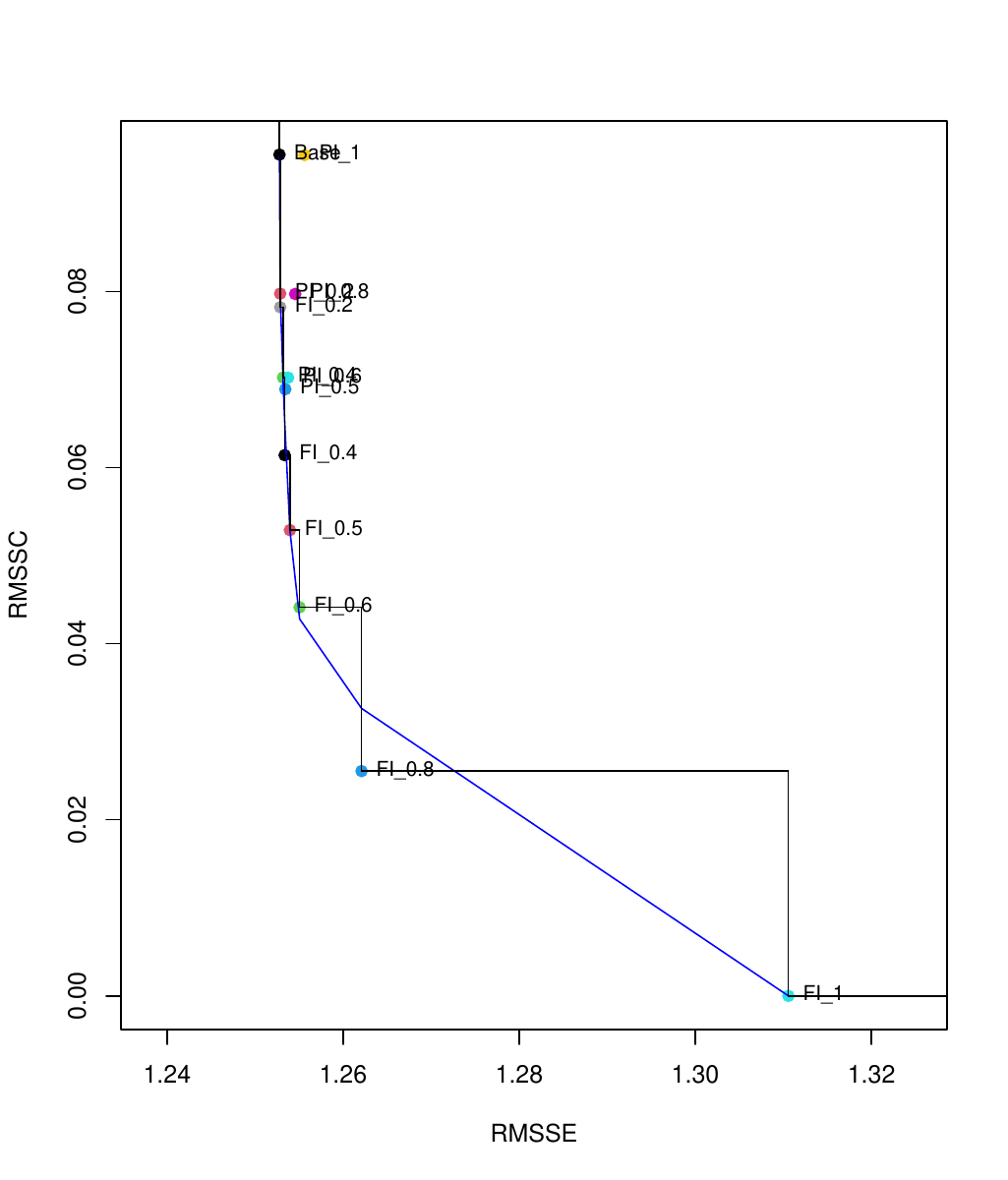}}
  \subfloat[ETS]{\includegraphics[height=0.4\textheight]{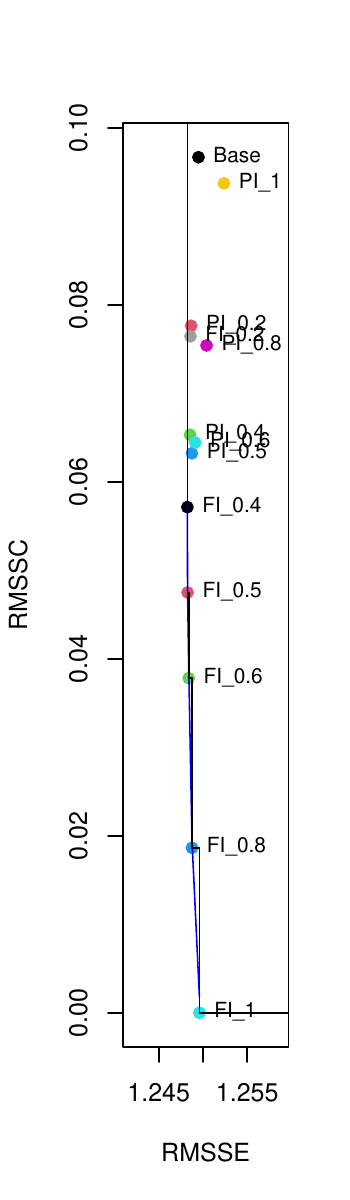}}
      \hfill
  \subfloat[ARIMA]{\includegraphics[height=0.4\textheight]{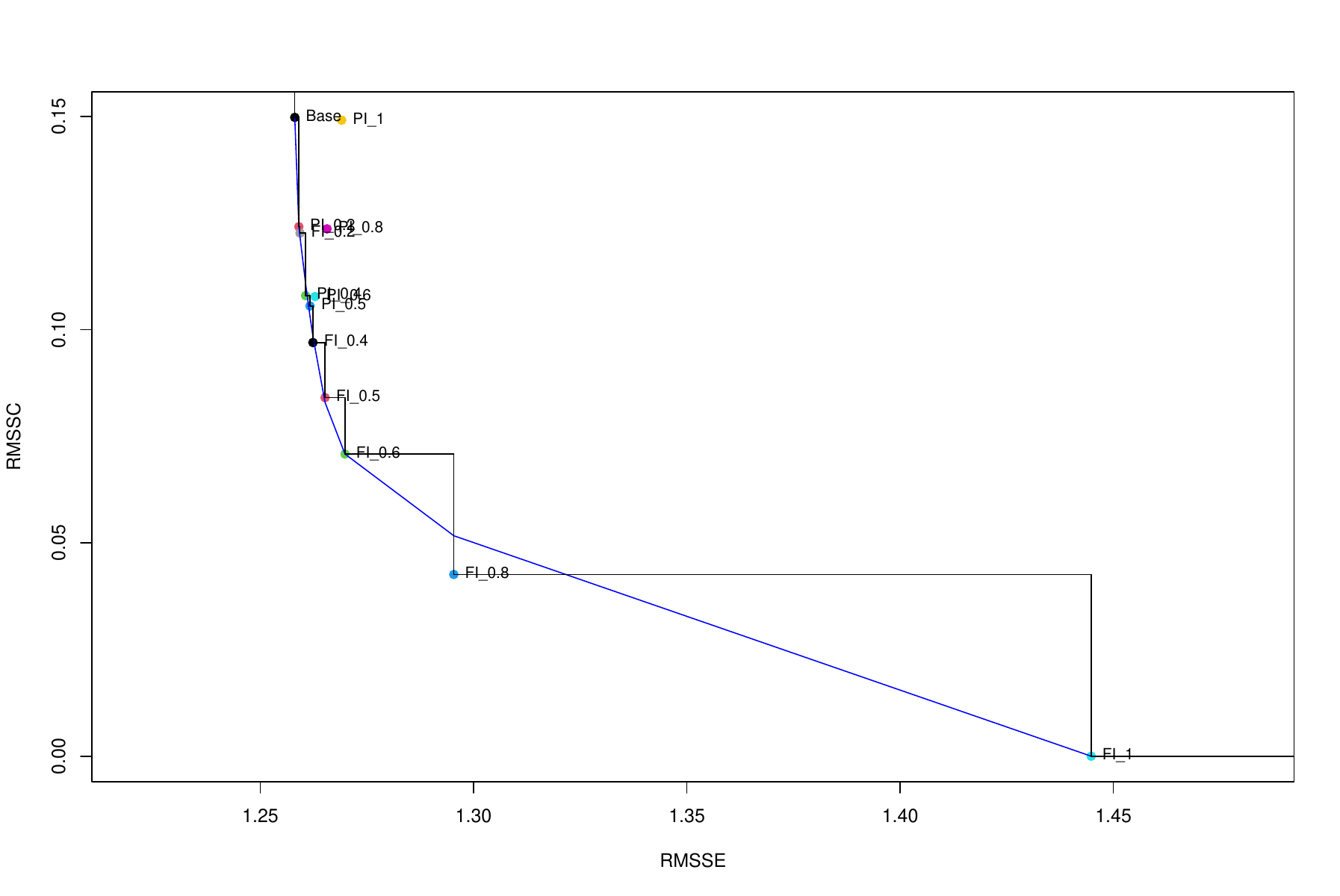}}
  \caption{Pareto fronts showing the accuracy (RMSSE) and stability (RMSSC) of all horizontal stability
experimental models across the M5 dataset. The blue lines show smoothed versions of the Pareto front with COBS splines, for cases where more than 3 solutions are on the Pareto front. The $x$ and $y$ axes use the same scale, so that a steep declining slope indicates a higher gain of stability (decrease of RMSSC) than what is lost in accuracy (increase of RMSSE).}
  \label{fig:h-M5}
\end{figure}

Figure~\ref{fig:v-M4} shows the Pareto fronts representing the accuracy (MASE) and stability (MASC) results across the M4 dataset for the vertical stability experiments. The plots are grouped according to the base model, and they use the same scale for $x$ and $y$ axes, so that a steep declining slope indicates a higher gain of stability (decrease in MASC) than what is lost in accuracy (increase in MASE). 
In the figure we see that for vertical stability there are no models that produce forecasts which are most accurate and stable at the same time with any base model. The Stable N-BEATS is not on the Pareto front, which is likely due to the fact that it is trained using RMSSE and the stability analogous measure RMSSC, while we report MASE and MASC. Using, e.g., sMAPE/sMAPC, its solution is on the Pareto front and thus, it is a good and valid method. However, it does not significantly stand out as there are other methods that are either more accurate or stable than the Stable N-BEATS. Hence, with our simpler interpolation approach, we can get comparable results. Furthermore, our method easily produces the full spectrum of the accuracy-stability trade-off whereas with the Stable N-BEATS approach, the users have to change the parameter, $\lambda$, and retrain the model multiple times to obtain solutions with different trade-offs which can be time-consuming. Presenting the full spectrum of the trade-off can be important for practitioners to select a model based on their requirements. For example, regarding both vertical and horizontal stability experiments on all base models across the M4 dataset, PI\_0.2 and FI\_0.2 can be selected to obtain more accurate forecasts and FI\_0.6 and FI\_0.8 can be selected to obtain more stable forecasts.

Figure~\ref{fig:h-M4} shows the results for horizontal stability on the M4, again using the same scale for $x$ and $y$ axes to facilitate the assessment of the trade-off between accuracy and stability. For N-BEATS and ETS we see results similar to the vertical stability, with an accurate base model and a trade-off with lower accuracy towards higher stability. However, for LightGBM, PR, and ARIMA, we see that very stable methods are also the most accurate, which is likely due to the methods failing to model the little information that may be left in the remainder of the seasonal decomposition on which the stabilisation is based.

Figures \ref{fig:v-M5} and \ref{fig:h-M5} show the Pareto fronts representing the accuracy (RMSSE) and stability (RMSSC) results across the M5 dataset for the vertical and horizontal stability experiments, respectively. The results differ from the Pareto fronts of the M4 dataset. We see that for vertical stability, the base model is never part of the Pareto front, which is especially clear for the PR and LightGBM models, where only very few solutions are on the Pareto front, indicating that on this dataset it is possible, for example with the variant FI\_0.8 for PR, to obtain a model that is both very accurate and very stable. We argue that this phenomenon also explains why the ensemble model proposed by \citet{IN20221386} won the M5 competition. The winning method also implicitly combines the forecasts obtained from different origins to produce the final forecasts and thus, the forecasts are stable and for this dataset, stable forecasts are also accurate. However, this does not work in the same way for other datasets and hence, presenting the full spectrum of the accuracy-stability trade-off is highly useful for decision-making in real-world applications.

The observations are slightly different in the horizontal stability experiments across the M5 dataset. As shown in Figure \ref{fig:h-M5}, the FI\_0.2 and FI\_0.6 models respectively provide the most accurate forecasts with the PR and LightGBM models across the M5 dataset. However, these variants do not provide the most horizontally stable forecasts. Thus, in this case practitioners can also select any model variant on the Pareto front based on their requirement to obtain forecasts. In general, the variant FI\_0.8 provides relatively accurate and horizontally stable forecasts with all models.
Enforcing further stability as does the FI\_1 solution leads to strong degradation in accuracy, especially for the PR and ARIMA models.
 In both Figures \ref{fig:v-M5} and \ref{fig:h-M5}, the base models are not on the Pareto fronts for all cases but one (horizontal stability of ARIMA) which highlights that linear interpolation can produce both accurate and stable forecasts compared with the base models for the M5 dataset.

While presenting the full Pareto front allows a practitioner to freely choose a solution that trades off stability and accuracy in the desired way, it is a legitimate question whether there are any methods or general guidelines to guide us in this trade-off or even fully automate it under certain assumptions.
If the Pareto front is convex (as it turns out to be in all our experiments) and both objectives are directly comparable and deemed equally important, we can use an elbow or curvature method and choose the point with the highest curvature or 2nd derivative from the Pareto front to achieve the best trade-off. 
The intuition here is that the point with the highest curvature marks the transition between solutions that gain significant stability with only a small loss in accuracy and solutions where further gains in stability come at the cost of a much larger loss in accuracy.
Without smoothing, the results can be relatively arbitrary. Thus, we smooth the Pareto front, for example, with a convex constrained B-spline \citep[COBS, ][]{he1999cobs,ng2007fast} to enforce convexity. 

For most practitioners, accuracy will be more important than stability, and often, the goal will be to achieve a solution that does not lose much accuracy (if any), but has greater stability.
In the literature, methods have been proposed to choose a particular solution from the Pareto front, taking into account the different importances of different objectives. An example is the R-method \citep{rao2021ranking}, which allows us to rank objectives according to their importance. 
In experiments not reported here in detail, the R-method always ranked the most accurate solution first, and the maximally stable solution second, across all our datasets. As both solutions would usually not be the desired solutions in terms of a desired trade-off, we therefore report in the following results from the method of using the 2nd derivative of the smoothed Pareto front, using a COBS spline, to determine the point with the best trade-off.

For brevity, we focus on the vertical stability case. Table~\ref{tab:pareto_vert} shows the results. 
We see that the method chooses solutions that always have small accuracy losses, typically under 2\%, and in two cases under 3\% and 5\%, respectively.
At the same time, the solutions have considerably higher stability, typically around 20-50\%, compared to the method that has the highest accuracy. Thus, this method yields good results in our experiments. To ensure that the accuracy does not drop too much, we could further implement a procedure such as the following. We define a maximal loss in accuracy $\delta_{max}$ that is acceptable, for example $\delta_{max}=3\%$. Then, if the solution obtained from the 2nd derivative method fulfills this constraint, we use this solution. If the solution is not sufficiently accurate, we choose the least accurate solution from the Pareto front that fulfills this accuracy constraint. With this procedure, we assume that each solution on the Pareto front before the solution with the highest curvature leads to less accuracy loss than it gains in stability compared with the previous solution on the Pareto front. This is typically the case.

\begin{table*}[ht]
\centering
\begingroup\footnotesize
\begin{tabular}{lrrrrrrr}
  \toprule
  & Method & MASE/ & MASC/ & $dec_{acc}$ (\%) & $inc_{stab}$ (\%) & $dec_{acc}$ & $inc_{stab}$ \\ 
  &  & RMSSE & RMSSC &  &  &  &  \\ 
\hline
 &  \multicolumn{7}{c}{M4} \\ 
\hline 
  N-BEATS & FI\_0.5 & 0.642 & 0.178 & 1.265 & 35.198 & 0.008 & 0.097 \\ 
  PR & FI\_0.6 & 0.829 & 0.115 & 4.802 & 47.243 & 0.038 & 0.103 \\ 
  LightGBM & FI\_0.5 & 0.878 & 0.174 & 2.780 & 38.275 & 0.024 & 0.108 \\ 
  ETS & FI\_0.4 & 0.663 & 0.190 & 1.152 & 27.968 & 0.008 & 0.074 \\ 
  ARIMA & PI\_0.4 & 0.641 & 0.191 & 1.184 & 30.670 & 0.008 & 0.084 \\ 
\hline 
 &  \multicolumn{7}{c}{M3} \\ 
\hline    
  N-BEATS & FI\_0.6 & 0.651 & 0.117 & 1.894 & 39.469 & 0.012 & 0.076 \\ 
  PR & FI\_0.2 & 0.760 & 0.156 & 0.621 & 17.806 & 0.005 & 0.034 \\ 
  LightGBM & FI\_0.2 & 0.770 & 0.193 & 0.412 & 19.381 & 0.003 & 0.046 \\ 
  ETS & FI\_0.4 & 0.621 & 0.133 & 0.695 & 36.180 & 0.004 & 0.076 \\ 
  ARIMA & FI\_0.4 & 0.622 & 0.131 & 0.648 & 36.311 & 0.004 & 0.075 \\ 
\hline 
 &  \multicolumn{7}{c}{Corporaci\'on Favorita} \\    
\hline    
  PR & FI\_0.6 & 0.585 & 0.043 & 0.082 & 31.167 & 0.000 & 0.020 \\ 
  LightGBM & FI\_0.2 & 0.633 & 0.077 & 0.212 & 18.685 & 0.001 & 0.018 \\ 
  ETS & FI\_0.5 & 0.567 & 0.030 & 0.071 & 30.992 & 0.000 & 0.013 \\ 
  ARIMA & FI\_0.8 & 0.573 & 0.015 & 0.205 & 55.217 & 0.001 & 0.018 \\ 
\hline 
 &  \multicolumn{7}{c}{M5} \\ 
\hline    
  PR & FI\_0.6 & 0.996 & 0.107 & 0.000 &  0.000 & 0.000 & 0.000 \\ 
  LightGBM & FI\_0.8 & 0.986 & 0.049 & 0.293 & 49.549 & 0.003 & 0.048 \\ 
  ETS & FI\_0.5 & 0.934 & 0.058 & 0.060 & 30.538 & 0.001 & 0.025 \\ 
  ARIMA & FI\_0.5 & 0.952 & 0.054 & 0.094 & 31.019 & 0.001 & 0.024 \\ 
   \bottomrule
\end{tabular}
\caption{Results of automatically choosing one solution from the Pareto front, for the vertical stability case. Accuracy decrease and stability increase are in comparison with the method that has the highest accuracy.}
\label{tab:pareto_vert}
\endgroup
\end{table*}

\section{Conclusions and Future Research}
\label{sec:conclusion}

Obtaining forecasts that are both stable and accurate is important for many real-world applications. In this paper, we have systematically explored different types of stability, and have proposed a categorisation based on same/different targets and origins, focusing then on the two types of different origin and same target, and same origin and different target, which we call vertical stability and horizontal stability. Making forecasts vertically stable across different forecast origins while maintaining accuracy is important for correct decision-making and planning. In addition, making forecasts horizontally stable across the forecast horizon can help reduce the bullwhip effect in supply chains. However, stable forecasting has received limited attention from the forecasting community. The available forecast stabilisation frameworks are only applicable to certain base models, and extending them to stabilise the forecasts provided by any base model is not straightforward. Furthermore, these frameworks are designed only to make forecasts vertically stable.

In this paper, we have proposed a simple model-agnostic linear interpolation approach to make forecasts either vertically or horizontally stable. To make the forecasts made at a given origin vertically stable, the forecasts are linearly combined with the corresponding forecasts made at the previous origin. To stabilise the forecasts horizontally across the forecast horizon, adjacent forecasts are linearly combined. To account for known and predictable non-stationarities in the horizontal case, the method can be applied to remainders after a seasonal decomposition. The decomposition can be part of the model (decompose-then-forecast), or can be an explicit step in the stabilisation process (forecast-then-decompose) to preserve the model-agnostic post-processing nature of  the procedure. 
The proportion of the previous forecasts used during the interpolation to make the current forecasts more stable is a parameter for the method that enables easy control of the trade-off between stability and accuracy. For both vertical and horizontal stability experiments, linear interpolation is conducted in two ways, partial interpolation and full interpolation. Experiments are conducted using various base models: N-BEATS, PR, LightGBM, ETS, and ARIMA. Across four experimental datasets, we have shown that our framework can produce considerably more stable forecasts with modest loss in accuracy compared to benchmark models on different error measures. Furthermore, our framework can produce forecasts that are both more accurate and stable than the benchmark models for some datasets across the error measures used. 

From our experiments, we conclude that linear interpolation is a good approach to make the forecasts either vertically or horizontally stable. We further conclude that full interpolation leads to more stable forecasts than partial interpolation with both vertical and horizontal stability types, as it considers the previously interpolated forecasts to make the next forecasts. 
For datasets with intermittent series such as Favorita and M5, the full interpolation models can provide both more accurate and stable forecasts than the base models. For the datasets with higher trends and seasonal effects, such as M3 and M4, the full interpolation models can provide forecasts that only lose small amounts of accuracy in exchange for being considerably more stable. If horizontal stabilisation is performed for non-stationary series on a remainder of a seasonal decomposition, stabilisation can reduce noise in the forecasts due to a low signal-to-noise ratio in the remainder.
In general, our approach enables practitioners to select a trade-off between accuracy and stability from the Pareto front, based on their requirements. Compared with other recently proposed forecast stability models, our interpolation-based framework is simple to implement and applicable to any base model to make forecasts either vertically or horizontally stable. Thus, we recommend using our full interpolation framework as a benchmark and an easy way to stabilise the forecasts, before potentially more sophisticated methods are tried.

There are several possible avenues for extending this research. Weighting mechanisms could be developed to change stability, in the case of horizontal stability with respect to trend, seasonality, holiday effects, known promotions, and other similar effects. In addition, different weightings for different horizons may be beneficial in certain situations. Finally, in the case of vertical stability, one point worthy of investigation is to develop measures of how much new information the newly available data add. Although this is a problem at the core of any forecasting method, how responsive it should be to the most recent observations is also relevant in a stability context, as truly new information may render stability less desirable.

\section*{Acknowledgements}

Christoph Bergmeir is supported by a María Zambrano (Senior) Fellowship that is funded by the Spanish Ministry of Universities and Next Generation funds from the European Union. The work was initiated while he was a short term employee at Meta Inc. We want to thank two anonymous reviewers whose comments have lead to considerable improvements of the paper. We have used Paperpal and ChatGPT-4o for language editing.

\bibliographystyle{elsarticle-harv}

\bibliography{references}

\end{document}